\documentclass[11pt]{article}

\PassOptionsToPackage{table}{xcolor}

\usepackage[preprint]{acl}

\usepackage{times}
\usepackage{latexsym}
\usepackage[T1]{fontenc}
\usepackage[utf8]{inputenc}
\usepackage{microtype}
\usepackage{inconsolata}
\usepackage{graphicx}

\usepackage{booktabs}
\usepackage{colortbl}
\definecolor{oursrow}{HTML}{F3EFFF}
\definecolor{oursink}{HTML}{6D28D9}
\definecolor{grouphead}{HTML}{4B5563}
\definecolor{groupband}{HTML}{F3F4F6}
\definecolor{keycol}{HTML}{F8F6FF}
\definecolor{gain}{HTML}{15803D}
\definecolor{drop}{HTML}{BE123C}
\newcommand{\up}[1]{\textsubscript{\textcolor{gain}{+#1}}}
\newcommand{\dn}[1]{\textsubscript{\textcolor{drop}{$-$#1}}}
\usepackage{amsmath}
\usepackage{amssymb}
\usepackage{amsfonts}
\usepackage{mathtools}
\usepackage{multirow}
\usepackage{makecell}
\usepackage{array}
\usepackage{enumitem}
\usepackage{nicefrac}
\usepackage{algorithm}
\usepackage{algorithmic}
\usepackage[capitalize,noabbrev]{cleveref}

\newcommand{\R}{\mathbb{R}}

\DeclareMathOperator*{\argmax}{arg\,max}

\DeclareMathOperator{\softmax}{softmax}

\newcommand{\vz}{\mathbf{z}}

\newcommand{\cL}{\mathcal{L}}

\newcommand{\lwm}{GUI-JEPA}
\newcommand{\enc}{f_{\theta}}
\newcommand{\pred}{g_{\phi}}
\newcommand{\zt}{\vz_t}
\newcommand{\zhat}{\tilde{\vz}_t} 

\newcommand{\cand}{c}
\newcommand{\cands}{\mathcal{C}_t}
\newcommand{\goal}{g}
\newcommand{\roi}{\operatorname{roi}}
\newcommand{\pool}{\operatorname{pool}}
\newcommand{\score}{s}
\newcommand{\ac}{AndroidControl}
\newcommand{\achigh}{AC-High}
\newcommand{\aclow}{AC-Low}

\graphicspath{{figures/}}

\title{GUI-JEPA: Candidate-Conditioned JEPA Latents\\as a Fast Action Policy for Mobile GUI Agents}

\author{
  Bingchao Wang\textsuperscript{1,2}\thanks{This work is completed during Bingchao Wang's research attachment at NTU.\quad\textsuperscript{\dag}Corresponding Authors.},
  Yongcheng Jing\textsuperscript{2},
  Shunyu Liu\textsuperscript{2},
  Chengyu Wang\textsuperscript{3},
  Jun Huang\textsuperscript{3},
  \\
  \bfseries Jie Yang\textsuperscript{1\dag},
  Wei Liu\textsuperscript{1\dag},
  Dacheng Tao\textsuperscript{2\dag}
  \\[2pt]
  \normalfont\normalsize
  \textsuperscript{1}Shanghai Jiao Tong University\quad
  \textsuperscript{2}Nanyang Technological University\quad
  \textsuperscript{3}Alibaba Cloud Computing
  \\
  \normalfont\normalsize
  \{bc\_wang, jieyang, weiliucv\}@sjtu.edu.cn,\ yongcheng.jing@ntu.edu.sg,
  \\
  \normalfont\normalsize
  \{shunyu.liu.cs, dacheng.tao\}@gmail.com,\ \{chengyu.wcy, huangjun.hj\}@alibaba-inc.com
}

\begin{document}
\maketitle
\begin{abstract}
Mobile GUI agents are typically fine-tuned 7B-scale vision-language models (VLMs) that generate each action as text.
However, these VLMs decode every action token by token, and even agents that score candidate actions keep a large language model or VLM in the loop, making them slow and memory-hungry.
To address these issues, we propose \lwm{}, to our knowledge the first decoder-free GUI agent policy built on a Joint-Embedding Predictive Architecture (JEPA), which selects actions in the latent space of a JEPA encoder pretrained on unlabelled GUI screenshots.
Its action- and region-conditioned latent predictor turns the screen latent into one latent per candidate action, letting lightweight heads score all candidates against the instruction in a single batched pass, without decoding a token.
We further show that the scorer relies on this latent pathway and that the predictor is trained for the decision itself: a randomly initialised predictor does just as well as one pretrained on screen dynamics.
On AndroidControl, \lwm{} is 2.8--4.0$\times$ faster on the same GPU than a Qwen2.5-VL-7B agent we fine-tune with the same inputs, while reaching 86--92\% of its in-distribution accuracy with 6\% of its parameters.
Code is available at \url{https://github.com/bcwang-sjtu/GUI-JEPA}.
\end{abstract}

\section{Introduction}
\label{sec:intro}

\begin{figure}[t]
  \centering
  \includegraphics[width=\columnwidth]{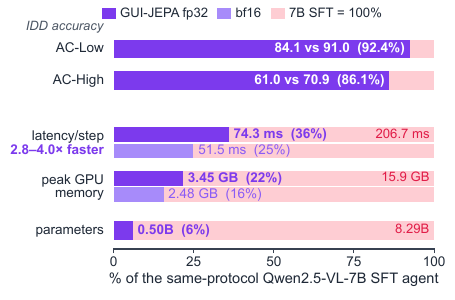}
  \caption{\textbf{Same inputs, same GPU: \lwm{} reaches 86--92\% of the 7B agent's accuracy at a fraction of its cost.} Bars are relative to our same-protocol Qwen2.5-VL-7B SFT agent, fine-tuned with LoRA (IDD \aclow{} / \achigh{} accuracy; latency: medians over 100 shared real \achigh{} steps on one H20 at batch 1; peak memory under HF serving; \cref{tab:latency}).}
  \label{fig:hero}
\end{figure}

A mobile graphical user interface (GUI) agent's per-step cost is dominated by a 7B+ vision-language model (VLM) decoding its action token by token. On one NVIDIA H20, a Qwen2.5-VL-7B agent \citep{bai2025qwen25vltechnicalreport} LoRA-fine-tuned \citep{DBLP:conf/iclr/HuSWALWWC22} on \ac{} needs 207\,ms per step under vLLM \citep{DBLP:conf/sosp/KwonLZ0ZY0ZS23} and 507\,ms under Hugging Face (HF) transformers to emit a JSON action of ${\sim}$12 tokens (median over 100 real steps, batch 1; \cref{tab:latency}), and 8B-class agents average \mbox{2.6--10.6\,s} per step in a live-device benchmark \citep{sui2026androiddailyverifiablebenchmarkmobile}. Yet in GUI grounding a VLM fixes its target elements during prefill, with decoding only refining coordinates \citep{lin2026happensdecodingprefilldetermines}, and decoding on the critical path makes agents correct but late \citep{dong2026guiagentscorrectlate}. Could a GUI agent \emph{score} its actions in the latent space of a joint-embedding predictive architecture \citep[JEPA;][]{assran2023self}, without decoding them at all?

GUI world models generate pixels, code or text and serve an LLM policy as simulator or verifier \citep{luo2026vimo,zheng2026code2worldguiworldmodel,DBLP:conf/iclr/MeiGCDLNJ26,cao2026mobiledreamergenerativesketchworld,xu2026mobileworldmodelguides}, or as a pretraining target \citep{liu2026scalingguiagentsvisual}. On GUIs, JEPAs have so far served understanding and verification: UI-JEPA describes user intent through an LLM decoder \citep{fu2025ui}, LGWM verifies GUI transitions \citep{zhang2026rightscreenwrongtransition}, and next-frame feature prediction regularises the training of a VLM agent \citep{yang2026mirage}. Candidate scorers over the accessibility tree (a11y) keep an LLM or VLM in the loop \citep{dai2026v,zhang2026jev,christianos2025lightweight}, and Agentic-JEPA, which acts with a JEPA, is text-only with 0\% out-of-distribution (OOD) success \citep{rodrigogines:hal-05546567}. To our knowledge, no GUI agent has yet used a JEPA as the policy that selects its actions.

\begin{figure*}[t]
  \centering
  \includegraphics[width=\textwidth]{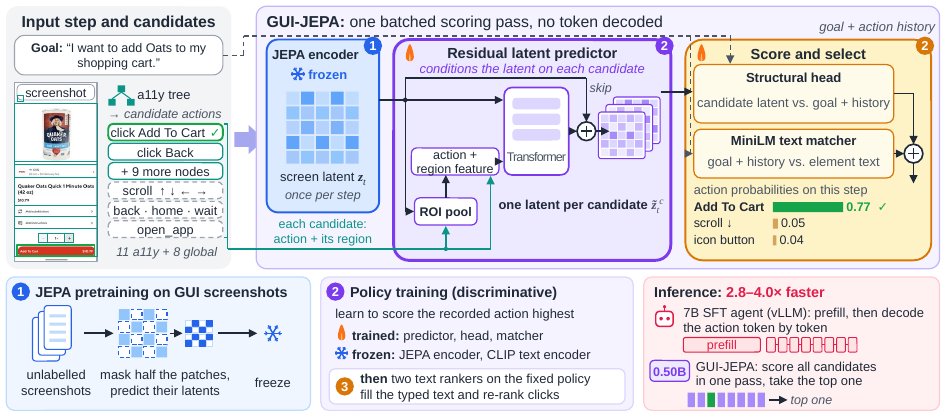}
\caption{\textbf{Overview of \lwm{}.} Top: on a real \ac{} step, the frozen JEPA encoder embeds the screenshot once, the residual predictor turns it into one latent per candidate in a single batch, and the structural head and the text matcher score every candidate (bars: a trained base policy's probabilities on this step; grids are schematic). Bottom: the training stages, and scoring in one pass versus decoding token by token (measured in \cref{sec:efficiency}). Circled numbers mark the stage that trains each module (1: JEPA pretraining, 2: policy training, 3: text rankers on the fixed policy); flame = trained, snowflake = frozen; green marks the recorded (correct) action.}
\label{fig:arch}
\end{figure*}

We bring JEPAs from perception and robotics to GUI control, moving action selection from the decoder into the latent space of a frozen JEPA encoder. \lwm{} encodes the screenshot once with a JEPA ViT-L pretrained on GUI screenshots, transforms that latent once per candidate (a11y element or global action) with a 39.9M residual predictor conditioned on the candidate's action and region of interest (ROI), scores the candidate-conditioned latents with a cosine structural head, adds a 33M text-matcher score and takes the argmax, all in one batched pass (\cref{fig:arch}). As in earlier candidate-scoring agents, the candidates come from the a11y tree, but \lwm{} never generates a token, so the cost of a step depends on how many candidates it scores.

Under the same inputs, judge and GPU as a Qwen2.5-VL-7B agent that we fine-tune with LoRA (SFT), \lwm{} is 2.8--4.0$\times$ faster and reaches 92.4\% / 86.1\% of its in-distribution (IDD) Low / High accuracy with 6.0\% of its parameters and 4.6--6.4$\times$ less GPU memory (\cref{fig:hero}).

Our contributions are:
\begin{itemize}
  \item \textbf{\lwm{}}, the first GUI agent whose policy is built on a JEPA. It selects actions by scoring candidate-conditioned JEPA latents, with no decoder, and is \mbox{2.8--4.0$\times$} faster than a same-protocol 7B agent that decodes token by token.
  \item \textbf{Adapting a JEPA policy to GUIs:} a11y candidates as actions, ROI- and action-conditioned residual prediction, a structural head whose only candidate-dependent visual input is the conditioned latent, and two text rankers that together add 3.8 out-of-distribution \achigh{} points (\cref{tab:waterfall}).
  \item \textbf{Mechanism:} the scorer relies on the JEPA latent pathway, which is trained for the decision. A randomly initialised predictor matches a dynamics-pretrained one, so a JEPA policy for GUIs needs no dynamics pretraining.
\end{itemize}

\section{Related Work}
\label{sec:related}

\paragraph{VLM GUI agents and their cost.}
The dominant mobile GUI agent is a VLM of 7B+ parameters that emits an action token by token from a screenshot \citep{DBLP:conf/iclr/0003WXWSJC0CL025,DBLP:conf/icml/XuWWLXSS0X25,qin2025ui,zhang2025agentcpm,bai2025qwen25vltechnicalreport}; 3B agents report \ac{} under incompatible protocols (\cref{tab:otherprotocol}). On-device agents need seconds per step \citep{park2026pastforwardfasterondevicegui}, and efficiency work cheapens the decoder \citep{bai2026efficientguiagentssystems}; \lwm{} removes it altogether.

\paragraph{Discriminative and candidate-scoring GUI agents.}
Closest in input are the a11y-text PaLM~2S baselines of \ac{} (\cref{sec:main_results}). LiMAC selects UI elements from screenshots and UI trees with a small action transformer, calls Florence-2 \citep{Xiao_2024_CVPR} for text arguments and reports its own metric \citep{christianos2025lightweight}; V-Droid verifies a11y candidates with a prefill-only LLM \citep{dai2026v}, Jev-Mobile selects a11y identifiers with a typed decision model \citep{zhang2026jev}, and IntentScore verifies a generative policy's actions \citep{chen2026intentscoreintentconditionedactionevaluation}; each keeps a generative LLM or VLM in the loop. \lwm{} shares a11y enumeration and argmax but scores a candidate-conditioned JEPA latent plus a 33M text matcher, with no generative model in the loop.

\paragraph{World models for GUI agents.}
GUI world models predict the next state as pixels \citep{luo2026vimo,ning2026imagineacthighfidelitydata}, code \citep{zheng2026code2worldguiworldmodel,xu2026appdeltaworldtransitiongroundeddeltacode}, text \citep{chae2025web,DBLP:conf/iclr/MeiGCDLNJ26} or textual element sketches \citep{cao2026mobiledreamergenerativesketchworld} and serve an LLM policy as simulator, verifier or data synthesiser; none of the four output modalities compared by \citet{xu2026mobileworldmodelguides} is joint-embedding. STP pretrains forward and inverse GUI dynamics in a unified multimodal model that still decodes actions \citep{liu2026scalingguiagentsvisual}, and \citet{li2026discriminativeworldmodelsweb} train a generative text world model over a11y-tree states by predicted-state matching, whose predictions feed a separate reward model that ranks candidate actions. \lwm{} is decoder-free and \emph{is} the policy, trained, like that web world model, for how it is used.

\paragraph{JEPA and latent world models as policies.}
In robotics, latent world models built on joint-embedding predictive architectures \citep{assran2023self,Assran2025VJEPA2S} plan \citep{DBLP:conf/icml/ZhouPLP25,maes2026leworldmodel,zholus2026robojepascalingroboticlatent} or drive action policies \citep{sun2026vla,lin2026jepawamlearningvisionlanguageactionpolicies,zhang2026vjepapolicybuildingeffective}. On GUIs, UI-JEPA feeds a JEPA-pretrained video encoder into a fine-tuned LLM decoder that describes user intent, with a reported 6.6$\times$ lower latency than Claude~3.5~Sonnet on IIW \citep{fu2025ui}; the concurrent LGWM, a JEPA over GUI transitions, verifies whether an observed transition is expected \citep{zhang2026rightscreenwrongtransition}, RSI-Anything's world-model predictions rank actions that a separate policy proposes \citep{cui2026rsianythingunlockingrecursiveselfimprovement}, and next-frame feature alignment serves as a training-time regulariser discarded at inference \citep{yang2026mirage}; the recent Agentic-JEPA picks actions by the cosine between predicted next-state and goal embeddings in synthetic text-only environments (100\% success in distribution, 0\% out of distribution) \citep{rodrigogines:hal-05546567}. Decision-aligned work finds that the latent closest to the goal need not be the best action \citep{liu2026djepadecisionalignedlatentworld} and that predictors can ignore the action \citep{gan2026actswmactionsensitiveworldmodels}; value-equivalent models likewise learn latents for planning without a reconstruction objective \citep{schrittwieser2020mastering}. In \lwm{} the JEPA is the policy itself: it scores a11y candidates through candidate-conditioned latents of a frozen GUI-pretrained encoder, with no decoder, whereas earlier GUI work uses a JEPA to describe intent, verify transitions or regularise training. Our mechanism result (\cref{sec:analysis_pathway}) adds a GUI-domain data point to work finding that predictability and action relevance come apart \citep{gu2026rightfutureactionlearning} and to work that trains world models for the decision \citep{li2026discriminativeworldmodelsweb}.

\section{Method: \lwm{}}
\label{sec:method}

\lwm{} chooses each action by scoring candidates. At every step it enumerates the executable actions from the a11y tree, embeds the screenshot once with a frozen JEPA encoder, transforms that screen latent once per candidate with a residual predictor, scores each candidate-conditioned latent and the candidate's text against the text condition, and executes the highest-scoring candidate (\cref{fig:arch}). Since no token, pixel or line of code is generated, a decision is one batched forward pass whose cost grows with the number of candidates.

\paragraph{Problem and notation.}
We learn a step-level offline policy on \ac{} \citep{li2024effects} that predicts the recorded action from one step's inputs and the ground-truth history. A step has a screenshot $x_t$, an a11y tree $T_t$, a text condition $\goal$ (the goal in \achigh{}, the step instruction in \aclow{}) and the ground-truth history $h_t$ (the previous action vector and the mean latent change from the previous frame; the text matcher receives the previous actions as text). A frozen encoder gives the screen latent, patch tokens $\zt=\enc(x_t)\in\R^{196\times1024}$. The candidate set $\cands=\mathrm{enumerate}(T_t)\cup\mathcal{G}$ joins element candidates (a click per enumerated node, an \texttt{input\_text} per editable one) with a fixed set $\mathcal{G}$ of global actions. A candidate $\cand$ carries a 16-d action vector $a_c$ (type one-hot plus coordinates or scroll direction) and, if it has an element, its region $\roi(c)$ (global actions: zero region feature, whole-screen pooling). The policy picks $\hat{c}_t=\argmax_{c\in\cands}\score(c)$.

\subsection{Input step and candidates}
\label{sec:candidates}
Element-level candidates are the visible a11y nodes flagged clickable, long-clickable, editable or scrollable, capped at the 80 smallest by area, which keeps specific elements and bounds the batch; each carries the text of the node and its descendants, its bounding box and its type. Every enumerated node yields a click candidate, and every editable node also an \texttt{input\_text} candidate whose text is filled in by the span ranker (\cref{sec:scoring}). Global actions act on the whole screen: the four scroll directions, \texttt{navigate\_back}, \texttt{navigate\_home}, \texttt{wait} and one \texttt{open\_app}, whose argument is the longest training-vocabulary app name in the text condition. The ground-truth element of a click is the smallest enumerated candidate that contains the recorded click point (details and coverage: \cref{app:protocol}).

\subsection{JEPA encoder}
\label{sec:encoder}
$\enc$ is a ViT-L/16 \citep[303M;][]{DBLP:conf/iclr/DosovitskiyB0WZ21}, ImageNet-initialised and pretrained (stage 1) with a JEPA objective adapted from I-JEPA \citep{assran2023self} on 189K unlabelled mobile (GUI-Odyssey; \citealp{Lu_2025_ICCV}) and desktop (VideoCUA; \citealp{jian2026cuasuitemassivehumanannotatedvideo}) screenshots, without action labels or \ac{} data. The student encoder sees the screenshot $\tilde{x}$ with a random half $M$ of its patches zeroed and regresses, at the masked positions, the tokens that an exponential-moving-average (EMA) teacher $\bar{f}$ computes on the full screenshot:
\begin{equation}
\label{eq:jepa}
\cL_{\mathrm{JEPA}}=\frac{1}{|M|}\sum_{i\in M}\ell_{\mathrm{sL1}}\big(\enc(\tilde{x})_i,\ \mathrm{sg}[\bar{f}(x)_i]\big),
\end{equation}
with $\ell_{\mathrm{sL1}}$ the smooth-$L_1$ loss and $\mathrm{sg}$ the stop-gradient. Random masking suits screenshots, whose content is spread over many small elements, and the latent target spares the encoder from reproducing fonts or colours that do not matter for acting. After two epochs the encoder is frozen and its patch tokens are cached, so policy training never runs the backbone (recipe and hyperparameters: \cref{app:implementation}; pretraining--test overlap: \cref{app:overlap}).

\subsection{Residual latent predictor}
\label{sec:predictor}
The predictor $\pred$ (39.9M) is a JEPA-style action-conditioned latent predictor that maps the screen latent and one candidate to a candidate-conditioned latent. Its action input concatenates $a_c$ with the ROI-pooled feature $\mathbf{u}_c=\pool_{\roi(c)}(\zt)\in\R^{1024}$ of the candidate's element (zero for global actions), so that two candidates of the same type, such as two buttons, are told apart by what their regions show as well as by their coordinates. An MLP of this pair is added to a learned query token, and a three-layer transformer attends over the 196 patch tokens and the query, giving the candidate-conditioned latent $\zhat^{c}=\zt+\Delta(\zt,a_c,\mathbf{u}_c)$. The residual form keeps the content of the current screen and lets the predictor express only a candidate-specific change, which preserves the detail needed to ground an element. All candidates of a step form one batch.
$\pred$ is initialised from a dynamics predictor trained on \ac{} transitions to predict the next step's latent from $\zt$ and the recorded action (\cref{app:mechanism}) and is then fine-tuned jointly with the scoring heads. Policy training adds no next-latent loss, and the policy loss alone shapes $\zhat^{c}$ into a decision feature (\cref{sec:analysis_pathway}).

\subsection{Score and select}
\label{sec:scoring}
\textbf{Structural head.}
The head pools the candidate-conditioned latent over the candidate's region (all 196 tokens for global actions) and over the whole screen, and compares an MLP of the pair by cosine with an MLP of the conditioning vector (learned scale $\kappa$, per-type bias):
\begin{equation}
\label{eq:score}
\begin{split}
\score_{\mathrm{lat}}(c)={}&\kappa\cos\!\big(\mathrm{MLP}_v(\mathbf{v}_c),\,\mathrm{MLP}_t(\mathrm{cond})\big)\\
&+b_{\mathrm{type}(c)},
\end{split}
\end{equation}
with $\mathbf{v}_c=[\pool_{\roi(c)}(\zhat^{c});\,\mathrm{mean}(\zhat^{c})]$.
$\mathrm{cond}$ joins the frozen CLIP embedding \citep[63M;][]{DBLP:conf/icml/RadfordKHRGASAM21} of $\goal$ with $h_t$. The head's only candidate-dependent visual input is $\zhat^{c}=\zt+\Delta$: $\zt$ reaches it via the residual skip and, independently of the candidate, through $\mathrm{cond}$. The constraint is deliberate. Everything the head sees of a candidate passes through the predictor, and training cannot bypass the JEPA latent pathway (\cref{sec:analysis_negatives}).

\textbf{MiniLM text matcher.}
The text matcher, a MiniLM-L12 cross-encoder \citep[33M;][]{DBLP:conf/nips/WangW0B0020,DBLP:conf/emnlp/ReimersG19}, scores $\goal$ plus the textual action history against the text of each element candidate ($\score_{\mathrm{text}}=0$ for global actions); the two scores are summed outside the structural head, $\score(c)=\score_{\mathrm{lat}}(c)+\score_{\mathrm{text}}(c)$. The structural head arbitrates the action type and the eight global actions, which have no text to match, and also contributes at element level, while the text matcher carries most of the grounding among clicks (\cref{sec:analysis_pathway}).

\textbf{Text rankers (stage 3).}
Two small rankers refine the arguments without changing which kind of action is taken. The \texttt{input\_text} argument is the top span of the goal-source span ranker, a MiniLM-L6 cross-encoder (22.7M) trained to rank spans of the goal, the text it reads at test time in \achigh{} (\cref{sec:waterfall}). If the argmax is a click, the type-gated click reranker $r$, a MiniLM-L12 listwise model (33.5M), re-scores the click candidates as $\score(c)+\beta\,\mathbf{1}[\mathrm{conf}(r)>\tau]\,\tilde r(c)$, with $\tilde r$ its $z$-scored logit and $\beta,\tau$ set on validation.

\subsection{Training}
\label{sec:training}
Training has three stages (circled numbers in \cref{fig:arch}): JEPA pretraining of the encoder (\cref{sec:encoder}), policy training, and the two text rankers. The base policy (predictor, structural head and text matcher, trained jointly; encoder and CLIP text encoder frozen) is trained on the 13{,}594 cached episodes (74{,}714 steps) of the 13{,}604-episode \ac{} training split with a listwise cross-entropy over the candidates of each step, $\cL_t=-w_t\log p_t(c^{*}_t)$ with $p_t=\softmax_{c\in\cands}\score(c)$, where $c^{*}_t$ is the ground-truth candidate and $w_t{=}1.5$ for non-click steps and 1 for clicks, which offsets the predominance of clicks. The objective matches inference, where the recorded action must outscore every alternative the policy could take on that screen. We train with AdamW for 6 epochs, one model for both \aclow{} and \achigh{} (hyperparameters: \cref{tab:hparams}), and fine-tune the predictor jointly, which outperforms freezing it (\cref{sec:analysis_negatives}). The two rankers are then trained on the fixed base policy. Of the 0.50B parameters, 366M are frozen (303M encoder, 63M CLIP text encoder) and ${\approx}$132M trained, including the 22.7M span ranker and the 33.5M click reranker.

\section{Experiments}
\label{sec:experiments}

\subsection{Setup}
\label{sec:setup}

\paragraph{Benchmark and splits.}
We test on the four official \ac{} subsplits \citep{li2024effects}: IDD (721 episodes) and the overlapping app\_unseen (631), task\_unseen (803) and category\_unseen (700, contained in task\_unseen). The full test set (1,543 episodes, 8,444 steps) and the OOD union (full test minus IDD, 4,586 steps) are the units of evidence. IDD results cover 716 of the 721 episodes (3,858 of 3,897 steps), because five IDD episodes also in OOD subsplits count as OOD (\cref{app:protocol}). The 137-episode validation split selected the click reranker's variant, weight $\beta$ and gate. Systems that state their split report all test subsets together, so we compare them with our full-test result and use IDD only against published IDD numbers.

\paragraph{Protocol.}
\aclow{} conditions on the step instruction and \achigh{} on the episode goal, both with the ground-truth action history. Our exact-match judge keeps the original benchmark's click rule (the chosen candidate's centre lies inside the ground-truth element box) and departs from the original protocol in four places: it counts no equivalent \texttt{open\_app} or navigation actions, requires the normalised typed text to match, accepts a click for a ground-truth \texttt{long\_press} (9 IDD steps), and takes the ground-truth box from the smallest enumerated candidate (originally the smallest a11y node), scoring clicks without a candidate (1.6--2.6\%) as errors where the original discards them (\cref{app:protocol}). Candidates come from the a11y tree, a privileged input that pixel-only agents do not use.

\subsection{Main results}
\label{sec:main_results}

\begin{table*}[t]
\centering\footnotesize
\setlength{\tabcolsep}{3.2pt}
\begin{tabular}{@{}llccccc@{}}
\toprule
System & Params & Input & Judge & Split & \aclow{} & \achigh{} \\
\midrule
\rowcolor{groupband}\multicolumn{7}{@{}l}{\color{grouphead}\emph{EM scores reported by other papers (screenshot input)}} \\
Aguvis-7B$^{a}$ & 7B & img & EM & unspec. & 89.4 & 54.2 \\
OS-Atlas-7B$^{a}$ & 7B & img & EM & unspec. & 67.3 & 56.5 \\
Qwen3-VL-8B$^{m}$ & 8B & img & EM$^{\ast}$ & all-split & 77.7 & 60.8 \\
Qwen2.5-VL-7B, released$^{a}$ & 7B & img & EM & unspec. & 85.0 & 62.9 \\
MIRAGE-4B$^{m}$ & 4B & img & EM$^{\ast}$ & all-split & 77.6 & 64.4 \\
Qwen2.5-VL-72B$^{q}$ & 72B & img & EM$^{\ast}$ & unspec. & 93.7 & 67.4 \\
AgentCPM-GUI-8B$^{a}$ & 8B & img & EM & unspec. & 90.2 & 69.2 \\
MIRAGE-8B$^{m}$ & 8B & img & EM$^{\ast}$ & all-split & 83.8 & 72.4 \\
UI-TARS-7B$^{a}$ & 7B & img & EM & unspec. & 91.8 & 74.4 \\
UI-Venus-Navi-7B$^{m}$ & 7B & img & EM$^{\ast}$ & all-split & 85.1 & 75.2 \\
\midrule
\rowcolor{groupband}\multicolumn{7}{@{}l}{\color{grouphead}\emph{Closest published protocol (a11y input, our click rule and split)}} \\
PaLM 2S, LoRA on a11y text$^{p}$ & n/a & a11y & relaxed & IDD & 83.2 & 70.8 \\
\midrule
\rowcolor{groupband}\multicolumn{7}{@{}l}{\color{grouphead}\emph{Same inputs, judge and split (ours)}} \\
Qwen2.5-VL-7B, LoRA SFT (our reference) & 8.29B & img+a11y & exact & IDD & 91.0 & 70.9 \\
\rowcolor{oursrow}\textbf{\lwm{} (ours)} & 0.50B & img+a11y & exact & IDD & 84.1 & 61.0 \\
\rowcolor{oursrow}\textbf{\lwm{} (ours)} & 0.50B & img+a11y & exact & full & 78.6 & 53.8 \\
\bottomrule
\end{tabular}
\caption{\textbf{\ac{} step accuracy under disclosed protocols.} \emph{img} = screenshot; \emph{a11y} = accessibility tree, a privileged input. Judges: \emph{exact} = ours (\cref{sec:setup}), \emph{EM} = exact match as reported ($^{\ast}$matching rule not defined in the source), \emph{relaxed} = the original \ac{} protocol. \emph{all-split} = MIRAGE's score over all subsplits; \emph{unspec.} = split not stated. Our 7B reference has one LoRA adapter per setting, same recipe (zero-shot: \cref{tab:sameprotocol}; latency: \cref{tab:latency}). Sources: $^{p}$\citet[Tab.~5]{li2024effects}; $^{a}$via \citet[Tab.~2]{zhang2025agentcpm} (its re-evaluations; ``released'': the released checkpoint); $^{m}$via \citet[Tab.~1]{yang2026mirage}; $^{q}$\citet[Tab.~9]{bai2025qwen25vltechnicalreport}. Other protocols: \cref{tab:otherprotocol,tab:otherprotocol2}.}
\label{tab:main}
\end{table*}

\paragraph{Closest published protocol.}
The \ac{} paper's LoRA-tuned PaLM~2S over a11y text \citep[Tab.~5]{li2024effects} shares our a11y input (we add the screenshot), click rule and subsplits, under the original relaxed judge. In distribution, \lwm{} matches it on \aclow{} (84.06 vs 83.2) and trails on \achigh{} (60.99 vs 70.8; \cref{tab:subsplit_waterfall,tab:subsplit_waterfall_low}).

\paragraph{Exact-match VLM agents.}
The pixel-only \mbox{7--8B} exact-match entries of \cref{tab:main} decode their actions with 14--16$\times$ more (nominal) parameters than \lwm{}. On \aclow{} our full-test 78.64 is above OS-Atlas-7B (67.3), MIRAGE-4B (77.6) and Qwen3-VL-8B (77.66), and none of these agents reports a comparable per-step latency.

\begin{figure*}[t]
  \centering
  \includegraphics[width=\textwidth]{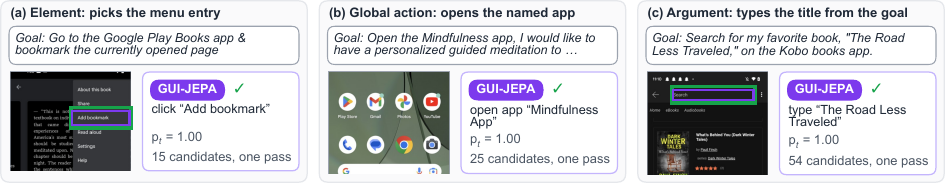}
\caption{\textbf{\lwm{} on three real \achigh{} IDD steps}, one per action kind: (a) an element click, (b) a global action, (c) a typed argument that the span ranker copies from the goal. Each step takes one scoring pass over all its candidates ($p_t$: probability of the chosen action; boxes: ground truth in green, \lwm{}'s choice in purple).}
\label{fig:qual_jepa}
\end{figure*}

\paragraph{Examples.}
\Cref{fig:qual_jepa} shows \lwm{} on three real IDD \achigh{} steps, one per action kind: it picks a menu entry among 15 candidates, opens the app named in the goal, and types a book title copied from the goal, each with one forward pass.

\subsection{Same-protocol 7B reference}
\label{sec:sameprotocol}

We also train a same-protocol reference: Qwen2.5-VL-7B-Instruct \citep{bai2025qwen25vltechnicalreport} fine-tuned with LoRA \citep{DBLP:conf/iclr/HuSWALWWC22} on 4,000 training episodes for one epoch, one adapter per setting (\lwm{}: one model, 13,594 episodes for six; 377-token image), given the same screenshot, candidates and ground-truth history as \lwm{} and scored by the same judge on the same 716 IDD episodes. It reaches 91.01 / 70.87 Low / High, and \lwm{} reaches 84.06 / 60.99, 92.4\% / 86.1\% of it, with 6.0\% of its parameters (0.498B vs 8.29B; 16.6$\times$ fewer) and 2.8--4.0$\times$ its vLLM speed (\cref{sec:efficiency}). The gap is 7.0 points when the step instruction is given and 9.9 with the goal alone. \Cref{sec:gap} breaks the \achigh{} gap down by action type.

\subsection{Gains from the two text rankers}
\label{sec:waterfall}

\begin{table}[t]
\centering\footnotesize
\setlength{\tabcolsep}{2.2pt}

\begin{tabular}{@{}lcccc@{}}
\toprule
Stage & IDD & full & OOD & val. \\
\midrule
Base policy & 56.84 & 49.79 & 43.85 & 57.54 \\
+ span ranker & 59.69 & 52.44 & 46.34 & 60.43 \\
\rowcolor{oursrow}+ click reranker & \textbf{60.99} & \textbf{53.75} & \textbf{47.67} & \textbf{62.46} \\
$\Delta$ total & \textcolor{gain}{\textbf{+4.15}} & \textcolor{gain}{\textbf{+3.96}} & \textcolor{gain}{\textbf{+3.82}} & \textcolor{gain}{\textbf{+4.92}} \\
\bottomrule
\end{tabular}
\caption{\textbf{Gains from the two text rankers} (\achigh{} step accuracy, \%; all stages on the same base-policy checkpoint; \aclow{}: \cref{tab:subsplit_waterfall_low}). OOD = full test minus IDD; val.\ = the 137-episode validation split, on which the reranker's variant, weight and gate were selected.}
\label{tab:waterfall}
\end{table}
\cref{tab:waterfall} isolates the two text rankers. Switching the training data of the 22.7M \texttt{input\_text} span ranker from step instructions to goal spans adds no parameters and +2.85 IDD High (+2.49 OOD union) by raising IDD High \texttt{input\_text} accuracy from 13.5 to 52.5. The type-gated click reranker then adds +1.30 IDD High (+1.33 OOD union) by re-ordering click candidates only. Together they lift OOD-union High from 43.85 to 47.67 and every unseen subsplit by 3.5--4.1 points (\cref{app:waterfall}), and the span-ranker gain replicates on the validation split (+2.89 vs +2.85 on IDD).

\subsection{Efficiency}
\label{sec:efficiency}\label{sec:cost}
\begin{table}[t]
\centering\footnotesize
\setlength{\tabcolsep}{2.2pt}
\begin{tabular}{@{}llrrrr@{}}
\toprule
System & Serving & p50 & p90 & Mem. & $\times$ ours \\
\midrule
7B SFT, merged & HF & 506.7 & 588.9 & 15.9 & 6.8/9.8 \\
7B SFT, merged & vLLM & 206.7 & 247.9 & $\ge$16 & \textbf{2.8}/\textbf{4.0} \\
\midrule
\rowcolor{oursrow}\textbf{\lwm{}}, fp32 & PyTorch & \textbf{74.3} & 127.8 & 3.45 & 1 \\
\rowcolor{oursrow}\textbf{\lwm{}}, bf16 & PyTorch & \textbf{51.5} & 79.7 & 2.48 & 1 \\
\bottomrule
\end{tabular}
\caption{\textbf{Same-hardware latency per real step} (ms; peak GPU memory in GB; same 100 \achigh{} IDD steps, one H20, batch 1). 7B = Qwen2.5-VL-7B (zero-shot and unmerged-LoRA timings: \cref{app:latency}); $\ge$: vLLM pre-allocates at least the weights. $\times$ ours: the 7B agent's p50 divided by our fp32 or bf16 p50.}
\label{tab:latency}
\end{table}

\begin{figure}[t]
  \centering
  \includegraphics[width=\columnwidth]{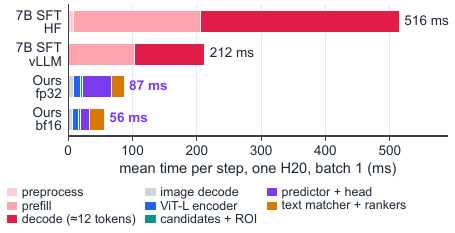}
  \caption{\textbf{Where a step's time goes} (segment means, 100 shared real steps, one H20, batch 1; medians: \cref{tab:latency}). Decoding takes half or more of the 7B step, and its prefill alone outlasts a whole \lwm{} step.}
  \label{fig:timebudget}
\end{figure}

\begin{figure*}[t]
  \centering
  \includegraphics[width=\textwidth]{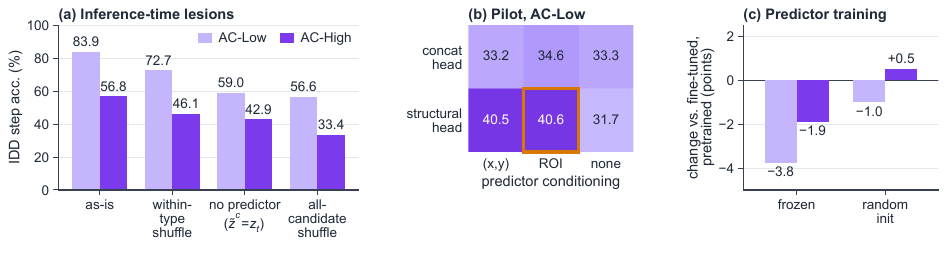}
  \caption{\textbf{The scorer relies on the latent pathway, which needs no dynamics pretraining.} (a)~Evaluation-time lesions (IDD; OOD and retrained controls in \cref{tab:lesion,tab:seeds_mech}). (b)~Pilot \aclow{} (an early configuration without the text matcher): only the structural head uses the pathway (orange box: deployed design). (c)~Change in IDD accuracy against the jointly fine-tuned, dynamics-pretrained predictor when the predictor is kept frozen or randomly initialised (identical recipe).}
  \label{fig:mechanism}
\end{figure*}

\paragraph{What the comparison fixes.}
The timing compares a scored step with a decoded step at fixed inputs, hardware and batch size. Both systems receive the same 100 real \achigh{} IDD screenshots, candidates and histories on one H20 at batch 1, and differ in model, training data and computation (\cref{sec:sameprotocol}). \lwm{}'s timing starts at JPEG decoding and covers CPU candidate and ROI preparation; the 7B agent's starts at image preprocessing and tokenisation. Screenshot capture, a11y retrieval and \lwm{}'s once-per-episode CLIP goal embedding (5.6\,ms) are excluded.

\paragraph{Speed.}
The final \lwm{} pipeline (base policy, span ranker and click reranker) needs a median 74\,ms per step in fp32 and 51.5\,ms with bf16 autocast, against 207\,ms for the 7B agent with vLLM and 507\,ms with HF transformers (merged LoRA; \cref{tab:latency}). \lwm{} is thus 2.8$\times$ (fp32) to 4.0$\times$ (bf16) faster than vLLM and 6.8--9.8$\times$ faster than HF, with 4.6--6.4$\times$ less peak GPU memory. In bf16 it makes the same predictions as in fp32 on all 200 timed steps, including the typed text (\cref{app:latency}).

\paragraph{Where the time goes.}
\Cref{fig:timebudget} splits the mean step time on the same 100 steps. Decoding takes 51\% (vLLM) to 60\% (HF) of the 7B agent's step, on top of its prefill, and the vLLM prefill alone outlasts a whole fp32 \lwm{} step, so the gain comes from both the smaller model and the absence of decoding. About half of a \lwm{} step goes to the batched predictor and head, which bf16 speeds up threefold, and its cost grows with the number of candidates (\cref{app:latency}).

\subsection{Error analysis}
\label{sec:gap}

Clicks, 63\% of the steps, account for 5.8 of the 9.9-point \achigh{} gap to the 7B agent, shared between choosing the click type and grounding the element, the part the text matcher carries (\cref{sec:scoring}). On global actions the gap lies mainly in the type decision, and on \texttt{open\_app} and \texttt{input\_text} in the argument (per type: \cref{tab:gap}). The two systems also err on different steps (\cref{fig:qual}): at least one of them is right on 78.7\% of IDD \achigh{} steps, and the margin between \lwm{}'s top two scores flags its errors (AUROC 0.74), so uncertain steps can be routed to a same-protocol 7B agent where accuracy matters more than speed (\cref{app:routing}).

\section{Analysis}
\label{sec:analysis}

\subsection{What the latent pathway does}
\label{sec:analysis_pathway}

\paragraph{Reliance.}
We lesion the trained base policy at evaluation time, text matcher untouched (\cref{fig:mechanism}a; \cref{tab:seeds_mech}). On IDD, shuffling the action vectors across all candidates costs 27.2 Low / 23.4 High, shuffling them within a type 11.2 / 10.8 (mostly scroll directions; about 5 High on clicks), and replacing the candidate-conditioned latent by the current latent (no predictor) 24.9 / 14.0, mostly on the global actions, which share the full-screen ROI. The action vector reaches the structural head only through the predictor, and training cannot route around it. Retrained at full scale with the identity map in place of the predictor, the policy reaches 62.36 / 42.33, 21.4 / 14.5 points below the base policy, again on the global actions, which the identity map leaves indistinguishable (\cref{tab:lesion}). The pathway thus carries chiefly action type and scroll direction, and also helps select among clicks.

\paragraph{The pathway is trained for the decision.}
A randomly initialised predictor trained with the identical full-scale recipe does as well as the dynamics-pretrained one (+0.50 High / $-1.01$ Low; \cref{fig:mechanism}c), and predictive auxiliary objectives (per-patch \citep{sun2026vla}, pooled, multi-horizon, LeWorldModel-style \citep{maes2026leworldmodel}) add nothing at 4K-episode scale (\cref{tab:lesion}). The predictor thus learns what the decision needs, and forward-model probes show that it forecasts the next latent worse than a copy of the current one (\cref{app:mechanism}). It acts as an action- and ROI-conditioned transform of the current screen, so a JEPA policy for GUIs needs no dynamics pretraining. This agrees with findings that predictability and action relevance come apart \citep{gu2026rightfutureactionlearning} and with training world models for the decision \citep{li2026discriminativeworldmodelsweb}.

\subsection{Design choices and the encoder}
\label{sec:analysis_negatives}\label{sec:analysis_encoder}

At pilot scale without the text matcher, the residual form raises \aclow{} from 40.6 to 43.9, whereas a predictor that re-synthesises the whole latent lowers \aclow{} click grounding to 34.8, below the identity map's 46.8. The structural head makes the policy use the predictor. Replacing the predictor by the identity map costs it 8.9 \aclow{} points (40.6 vs 31.7), whereas a concat-MLP head that also sees raw candidate features stays within 1.4 points with or without a predictor (\cref{fig:mechanism}b). Conditioning on the previous action and frame change adds 8 High points; at full scale, joint fine-tuning beats a frozen predictor by 3.8 \aclow{} / 1.9 \achigh{} IDD points (\cref{fig:mechanism}c; alternatives: \cref{tab:negatives}). The encoder is cheap to obtain. After 0.76 GPU-hours of JEPA pretraining on unlabelled screenshots, its frozen tokens score 67.22 on a coarse localisation probe over ScreenSpot-Pro \citep{Li2025ScreenSpotProGG} (probe hit@8; random-init ViT-L: 58.8), on par with DINOv3-ViT-L \citep{simeoni2026dinov3} (65.28; \cref{fig:encoder}, \cref{app:encoder}). In the policy, DINOv2-L \citep{oquab2024dinov2learningrobustvisual} features at 518\,px instead cost 17.4 / 9.8 points at 4K-episode scale, though it also changes the input resolution (\cref{tab:negatives}).

\section{Conclusion}
\label{sec:conclusion}

\lwm{}, to our knowledge the first GUI agent whose policy is built on a JEPA, replaces action decoding with one batched scoring pass over candidate-conditioned JEPA latents. With 0.50B parameters it needs 51.5--74\,ms per step, \mbox{2.8--4.0$\times$} less than a vLLM-served 7B agent on the same inputs, and reaches 92.4\% / 86.1\% of its IDD Low / High accuracy with 6.0\% of its parameters.

The JEPA latent pathway is essential to the scorer (lesions cost 10.8--23.4 IDD High points) and is trained for the decision: a randomly initialised predictor does as well as a dynamics-pretrained one. A JEPA used as a GUI action scorer should thus be judged by the decisions it supports, not by next-latent fidelity, and can be trained for them directly. Extending such scorers from steps to tasks is the next step for JEPA-based GUI agents.

\section*{Limitations}
\lwm{} consumes the a11y tree, a privileged structured input that pixel-only agents do not use; the 1.6--2.6\% of test clicks without a candidate count as errors, and every latency excludes screen capture and a11y retrieval. The policy is a step-level scorer; online task completion, which needs a termination head and a planner, is left to future work, as is using the predictor as a forward model. Design choices other than the reranker's variant, weight and gate were made on test IDD and re-checked on validation only for the span ranker and overall base-policy accuracy (\cref{sec:waterfall}). Speed-ups are measured at batch 1 against vLLM and HF serving of the 7B agent (\cref{sec:efficiency}). The same-protocol 7B reference is one LoRA adapter per setting trained on 4,000 of 13,594 episodes. The method is specialised to mobile a11y candidates; other candidate sources, such as web DOM nodes, need their own candidate texts and action vectors. Non-JEPA encoders are compared through a probe on the frozen tokens (\cref{app:encoder}); a comparison inside the full-scale policy is future work.

\section*{Ethical Considerations}
\lwm{} lowers the cost of GUI agents that act on a user's device. A wrong action, such as sending a message, deleting data or confirming a payment, can be hard to undo, and step-level errors remain common; a deployment needs user confirmation before irreversible actions. The speed that makes on-device agents practical also makes it cheaper to automate interactions meant for humans, such as mass account creation. The policy reads the accessibility tree, which can contain personal on-screen data; running the 0.50B model on-device keeps that data local. We use only public datasets under their licences (\ac{} Apache-2.0, GUI-Odyssey CC-BY-4.0, VideoCUA and ScreenSpot-Pro MIT), collect no new data, and involve no human subjects, annotators or crowd workers.

\bibliography{references}

\appendix
\setcounter{topnumber}{3}\setcounter{dbltopnumber}{3}\setcounter{totalnumber}{5}
\renewcommand{\topfraction}{0.9}\renewcommand{\dbltopfraction}{0.9}\renewcommand{\textfraction}{0.08}
\renewcommand{\floatpagefraction}{0.75}\renewcommand{\dblfloatpagefraction}{0.75}
\section*{Appendix Contents}
{\small
\begin{description}[leftmargin=1.4em,labelwidth=1.0em,labelsep=0.4em,itemsep=1pt,parsep=0pt,topsep=2pt,font=\normalfont\bfseries]
  \item[\ref{app:repro}] \textbf{Reproducibility statement}\dotfill\pageref{app:repro}
  \item[\ref{app:protocol}] \textbf{Protocol details}\dotfill\pageref{app:protocol}
  \begin{description}[leftmargin=2.4em,labelwidth=2.0em,labelsep=0.4em,itemsep=0pt,parsep=0pt,topsep=0pt,font=\normalfont]
    \item[\ref{app:seeds}] Per-seed results\dotfill\pageref{app:seeds}
    \item[\ref{app:bootstrap}] Episode-clustered bootstrap intervals\dotfill\pageref{app:bootstrap}
  \end{description}
  \item[\ref{app:waterfall}] \textbf{Per-subsplit waterfall}\dotfill\pageref{app:waterfall}
  \item[\ref{app:mechanism}] \textbf{Mechanism analyses}\dotfill\pageref{app:mechanism}
  \item[\ref{app:negatives}] \textbf{Design alternatives not adopted}\dotfill\pageref{app:negatives}
  \item[\ref{app:baselines}] \textbf{Other-protocol baselines}\dotfill\pageref{app:baselines}
  \item[\ref{app:encoder}] \textbf{Encoder probe}\dotfill\pageref{app:encoder}
  \begin{description}[leftmargin=2.4em,labelwidth=2.0em,labelsep=0.4em,itemsep=0pt,parsep=0pt,topsep=0pt,font=\normalfont]
    \item[\ref{app:overlap}] Pretraining--test near-duplicates\dotfill\pageref{app:overlap}
  \end{description}
  \item[\ref{app:implementation}] \textbf{Implementation details and hyperparameters}\dotfill\pageref{app:implementation}
  \item[\ref{app:latency}] \textbf{Latency measurement details}\dotfill\pageref{app:latency}
  \item[\ref{app:routing}] \textbf{Confidence routing to the 7B agent}\dotfill\pageref{app:routing}
\end{description}}

\section{Reproducibility statement}
\label{app:repro}
The encoder is a ViT-L/16 at 224\,px, initialised from ImageNet weights and trained with a JEPA objective adapted from I-JEPA \citep{assran2023self} on 189,262 unlabelled screenshots from GUI-Odyssey \citep{Lu_2025_ICCV} and VideoCUA \citep{jian2026cuasuitemassivehumanannotatedvideo} for 2 epochs ($\approx$0.76 GPU-hour on one H20), with no action labels and no \ac{} data (\cref{sec:encoder,app:implementation}). The policy is trained on the 13,594 cached episodes of the 13,604-episode \ac{} training split \citep{li2024effects} with a listwise cross-entropy, AdamW and 6 epochs, for three seeds (\cref{sec:training,tab:hparams}). We evaluate on the official test subsplits with the exact-match judge of \cref{sec:setup} and time both systems on the same 100 real IDD steps on one H20 at batch 1 (\cref{app:latency}).

\section{Protocol details}
\label{app:protocol}

\paragraph{Splits and overlap.}
The official \texttt{test\_subsplits.json} of \ac{} \citep{li2024effects} lists 721 IDD, 631 app\_unseen, 803 task\_unseen and 700 category\_unseen episodes, whose union is the full test set of 1,543 episodes and 8,444 steps. The subsplits are not disjoint: every category\_unseen episode is also a task\_unseen episode, and app\_unseen shares 606 episodes with category\_unseen and 607 with task\_unseen. We therefore treat the full test set and the OOD union (full test minus IDD as evaluated: 827 episodes, 4,586 steps) as the units of evidence and report the three OOD subsplits, each evaluated in full, only as a breakdown (\cref{app:waterfall}). Five of the 721 IDD episodes also appear in OOD subsplits and count as OOD, so every IDD accuracy in the paper is computed on 716 episodes and 3,858 of the 3,897 steps (99.0\%); the coverage analysis below uses all 721 episodes. The 137-episode validation split (690 steps) selected the click reranker's variant, its weight $\beta$ and its confidence gate. Every other design choice, including both text rankers, was made on test IDD; the validation row of \cref{tab:subsplit_waterfall} checks how these choices carry over. We train on 13,594 of the 13,604 training episodes of \citet[Tab.~3]{li2024effects} (74,714 of 74,722 steps).

\paragraph{Candidate enumeration and coverage.}
Element-level candidates are the visible accessibility-tree (a11y) nodes flagged clickable, long-clickable, editable or scrollable, plus containers that inherit clickability, de-duplicated by bounds and capped at the 80 smallest by area; editable nodes also yield \texttt{input\_text} candidates. Global actions are the four scroll directions, \texttt{navigate\_back}, \texttt{navigate\_home}, \texttt{wait} and one \texttt{open\_app}, whose argument is the longest training-vocabulary app name in the goal or instruction. A ground-truth click or long-press is mapped to the smallest enumerated candidate that contains its coordinate, whereas the original benchmark takes the smallest a11y node of any kind; the click rule itself (the chosen candidate's centre inside the ground-truth box) is unchanged. When no candidate contains the coordinate the step is unmapped and counts as an error for our policy (the original benchmark discards such steps). The mapped fraction is 97.4--98.4\% on the test subsplits (IDD 98.0\%, 2,405 of 2,454 click and long-press steps; full test 97.6\%, 4,961 of 5,083), 98.8\% on validation and 98.4\% on the training split. An IDD step has 23.6 a11y candidates on average (median 18, p95 57; 22.2 on the full test set and 23.2 on the training split), and the 80-candidate cap is reached on 16 IDD, 22 full-test and 272 training steps. Per-type IDD step counts are in \cref{tab:gap}.

\paragraph{Judge.}
A step is correct under our exact-match judge when the predicted action type equals the ground-truth type and the argument matches: a click when the centre of the chosen candidate lies inside the ground-truth element box, which is the click rule of the original benchmark \citep[App.~D.3]{li2024effects}; an \texttt{input\_text} when the typed string equals the ground-truth string after normalisation; \texttt{open\_app} and scroll when the argument matches exactly. The judge departs from the original protocol in four places, in both directions. (i)~It is stricter in accepting no equivalent actions, whereas the original protocol treats \texttt{open\_app} and a click on an element bearing the app name as equivalent, and likewise \texttt{navigate\_back} and a click on the on-screen Back button. (ii)~It is more lenient on \texttt{input\_text}: it compares the normalised text only, whereas the original protocol scores typing as an element action that also requires the target element to lie inside the ground-truth box and the text to be identical. (iii)~It is more lenient in accepting a click for a ground-truth \texttt{long\_press} (9 IDD steps). (iv)~It takes the ground-truth box from our enumerated candidates: a mapped click is judged against a box at least as large as the original's smallest a11y node (more lenient), while an unmapped click counts as an error, where \citet[App.~D.1]{li2024effects} discard it (stricter). On click steps, 62.7\% of IDD steps, both judges thus apply the same centre-in-box rule. Exact-match (EM) re-evaluations in the literature are the closest published protocol; numbers under the original relaxed protocol are not interchangeable with ours.

\paragraph{Conditioning and seeds.}
One trained model serves both settings: \aclow{} conditions on the step instruction and \achigh{} on the episode goal; both receive the ground-truth action history. We train three policy seeds. The main text reports seed 3, the best seed on every test split, at every stage and in both settings; full-scale results in this appendix are means $\pm$ sd over the three seeds unless a seed is named, and \cref{tab:seeds,tab:seeds_mech} list every seed. For $n{=}3$ the 95\% $t$-interval half-width, $4.30\,\mathrm{sd}/\sqrt{3}$, is about $\pm$2.4 points (IDD High $\pm$2.37, full-test High $\pm$2.34, IDD Low $\pm$2.52). Seeds capture training variance only, so external comparisons also use episode-clustered bootstrap intervals (\cref{app:bootstrap}).

\subsection{Per-seed results}
\label{app:seeds}
\begin{table*}[t]
\centering\footnotesize
\setlength{\tabcolsep}{3.2pt}
\begin{tabular}{@{}llcccccccc@{}}
\toprule
 & & \multicolumn{4}{c}{\achigh{}} & \multicolumn{4}{c}{\aclow{}} \\
\cmidrule(lr){3-6}\cmidrule(lr){7-10}
Split & Stage & seed 1 & seed 2 & seed 3 & mean $\pm$ sd & seed 1 & seed 2 & seed 3 & mean $\pm$ sd \\
\midrule
IDD & Base policy & 56.35 & 55.21 & \textbf{56.84} & 56.13$\pm$0.84 & 82.71 & 81.88 & \textbf{83.72} & 82.77$\pm$0.92 \\
 & + span ranker & 59.49 & 58.24 & \textbf{59.69} & 59.14$\pm$0.79 & 82.63 & 81.78 & \textbf{83.64} & 82.68$\pm$0.93 \\
 & + click reranker & 60.26 & 59.10 & \textbf{60.99} & 60.12$\pm$0.95 & 83.18 & 82.04 & \textbf{84.06} & 83.09$\pm$1.01 \\
\midrule
Full test & Base policy & 49.05 & 48.51 & \textbf{49.79} & 49.12$\pm$0.64 & 78.13 & 76.65 & \textbf{78.74} & 77.84$\pm$1.07 \\
 & + span ranker & 51.79 & 51.21 & \textbf{52.44} & 51.81$\pm$0.62 & 77.43 & 75.99 & \textbf{78.06} & 77.16$\pm$1.06 \\
 & + click reranker & 52.93 & 51.87 & \textbf{53.75} & 52.85$\pm$0.94 & 78.01 & 76.26 & \textbf{78.64} & 77.64$\pm$1.23 \\
\midrule
OOD union & Base policy & 42.91 & 42.87 & \textbf{43.85} & 43.21$\pm$0.55 & 74.27 & 72.24 & \textbf{74.55} & 73.69$\pm$1.26 \\
 & + span ranker & 45.31 & 45.29 & \textbf{46.34} & 45.65$\pm$0.60 & 73.05 & 71.13 & \textbf{73.35} & 72.51$\pm$1.21 \\
 & + click reranker & 46.75 & 45.79 & \textbf{47.67} & 46.74$\pm$0.94 & 73.66 & 71.39 & \textbf{74.07} & 73.04$\pm$1.44 \\
\midrule
app\_unseen & Base policy & 41.29 & 41.04 & \textbf{41.58} & 41.30$\pm$0.27 & 73.55 & 71.57 & \textbf{73.84} & 72.99$\pm$1.24 \\
 & + span ranker & 43.37 & 43.14 & \textbf{43.65} & 43.39$\pm$0.26 & 72.32 & 70.33 & \textbf{72.60} & 71.75$\pm$1.24 \\
 & + click reranker & 44.95 & 43.60 & \textbf{45.12} & 44.56$\pm$0.83 & 73.12 & 70.56 & \textbf{73.44} & 72.37$\pm$1.58 \\
\midrule
task\_unseen & Base policy & 42.63 & 42.61 & \textbf{43.84} & 43.03$\pm$0.70 & 74.19 & 72.22 & \textbf{74.57} & 73.66$\pm$1.26 \\
 & + span ranker & 45.39 & 45.41 & \textbf{46.57} & 45.79$\pm$0.68 & 73.16 & 71.24 & \textbf{73.54} & 72.65$\pm$1.23 \\
 & + click reranker & 46.82 & 45.90 & \textbf{47.89} & 46.87$\pm$1.00 & 73.77 & 71.48 & \textbf{74.19} & 73.15$\pm$1.46 \\
\midrule
category\_unseen & Base policy & 41.40 & 41.51 & \textbf{42.23} & 41.71$\pm$0.45 & 73.45 & 71.19 & \textbf{73.89} & 72.84$\pm$1.45 \\
 & + span ranker & 43.74 & 43.87 & \textbf{44.56} & 44.06$\pm$0.44 & 72.35 & 70.11 & \textbf{72.78} & 71.75$\pm$1.43 \\
 & + click reranker & 45.28 & 44.33 & \textbf{45.98} & 45.20$\pm$0.83 & 73.04 & 70.37 & \textbf{73.48} & 72.30$\pm$1.68 \\
\midrule
Validation & Base policy & 58.99 & 55.65 & \textbf{57.54} & 57.39$\pm$1.67 & 84.64 & 83.04 & \textbf{84.06} & 83.91$\pm$0.81 \\
 & + span ranker & 62.17 & 58.70 & \textbf{60.43} & 60.43$\pm$1.74 & 84.20 & 82.61 & \textbf{83.62} & 83.48$\pm$0.80 \\
 & + click reranker & 62.61 & 58.84 & \textbf{62.46} & 61.30$\pm$2.13 & 84.93 & 83.19 & \textbf{84.35} & 84.16$\pm$0.89 \\
\bottomrule
\end{tabular}
\caption{\textbf{Per-seed step accuracy of the three policy seeds} at every stage of the waterfall (exact match; bold: seed 3, reported in the main text). Stages share the base checkpoint of each seed; the span ranker is trained once and shared by the three seeds, the click reranker once per seed. The OOD union is derived per seed from the full-test and IDD step counts.}
\label{tab:seeds}
\end{table*}

\begin{table*}[t]
\centering\footnotesize
\setlength{\tabcolsep}{5pt}
\begin{tabular}{@{}lcccccc@{}}
\toprule
 & \multicolumn{3}{c}{IDD \aclow{}} & \multicolumn{3}{c}{IDD \achigh{}} \\
\cmidrule(lr){2-4}\cmidrule(lr){5-7}
Condition & s1 & s2 & s3 & s1 & s2 & s3 \\
\midrule
\multicolumn{7}{@{}l}{\color{grouphead}\emph{Inference-time lesions (text matcher untouched unless stated)}} \\
Base policy (as is) & 82.79 & 81.80 & 83.88 & 56.48 & 55.39 & 56.84 \\
\quad no predictor ($\zhat^{c}:=\zt$) & 61.79 & 62.05 & 59.02 & 41.16 & 40.93 & 42.87 \\
\quad within-type action shuffle & 72.01 & 71.12 & 72.65 & 46.47 & 44.95 & 46.06 \\
\quad all-candidate action shuffle & 55.83 & 55.81 & 56.64 & 35.17 & 34.63 & 33.41 \\
\quad swapped, non-fine-tuned predictor & 68.33 & 66.77 & 64.54 & 46.03 & 45.41 & 45.08 \\
\quad text matcher dropped & 39.17 & 36.39 & 39.48 & 30.82 & 29.32 & 31.60 \\
\midrule\multicolumn{7}{@{}l}{\color{grouphead}\emph{Training-time variants (identical full-scale recipe)}} \\
Jointly fine-tuned predictor (base) & 82.71 & 81.88 & 83.72 & 56.35 & 55.21 & 56.84 \\
Frozen dynamics-pretrained predictor & 80.43 & 80.38 & 79.94 & 54.90 & 54.80 & 54.95 \\
Randomly initialised predictor, joint & 82.84 & 82.24 & 82.71 & 57.52 & 56.14 & 57.34 \\
Identity map in place of the predictor & 61.53 & 62.96 & 62.36 & 42.79 & 41.21 & 42.33 \\
Matcher only (no latent pathway) & 55.60 & 54.98 & 55.42 & 40.51 & 39.17 & 40.10 \\
No predictor, action vector into a param.-matched head & 82.37 & 81.96 & 53.89 & 56.45 & 56.82 & 41.08 \\
\bottomrule
\end{tabular}
\caption{\textbf{Per-seed IDD accuracy of the mechanism controls} (s1--s3: training seeds; the main text reports seed 3). Lesion rows apply each intervention to the trained base policies. Means and out-of-distribution cells: \cref{tab:lesion}.}
\label{tab:seeds_mech}
\end{table*}

\Cref{tab:seeds,tab:seeds_mech} list every seed; the main-text values are the seed-3 columns, which lie 0.87 / 0.90 \achigh{} points above the three-seed means on IDD / the full test. Out-of-distribution lesion costs of the main-text model are given in \cref{app:mechanism}.

\subsection{Episode-clustered bootstrap intervals}
\label{app:bootstrap}
Because steps of one episode are correlated, the bootstrap for the external comparisons (\cref{sec:setup}, \cref{tab:main}) resamples test episodes. Each of $B{=}2{,}000$ resamples draws the 716 IDD (or 1,543 full-test) episodes with replacement and keeps every step of a drawn episode; the same resample is applied to the per-step predictions of the three seeds of the final system, whose step accuracies are averaged, and the 2.5th and 97.5th percentiles of this seed-averaged accuracy form the 95\% interval (the per-step predictions reproduce the reported accuracies exactly). The intervals are 60.12 [58.45, 61.85] High and 83.09 [81.72, 84.41] Low on IDD, and 52.85 [51.59, 54.04] High and 77.63 [76.62, 78.63] Low on the full test set (77.64 in \cref{tab:seeds} averages per-seed results rounded to two decimals). For seed 3 alone they are 60.99 [58.95, 62.97] High and 84.06 [82.46, 85.47] Low on IDD, and 53.75 [52.40, 55.06] High and 78.64 [77.61, 79.70] Low on the full test set. The bootstrap covers test-episode sampling and the $t$-interval over seeds covers training variance; neither covers the protocol differences listed in \cref{tab:main}.

\section{Per-subsplit waterfall}
\label{app:waterfall}

\begin{table*}[t]
\centering\footnotesize
\setlength{\tabcolsep}{4pt}

\begin{tabular}{llccccc}
\toprule
Subsplit & Episodes / steps & Base & + ranker & + reranker & $\Delta$ & PaLM 2S \\
\midrule
IDD & 716 / 3,858 & 56.13$\pm$0.84 & 59.14$\pm$0.79 & 60.12$\pm$0.95 & +3.98 & 70.8 \\
app\_unseen & 631 / 3,475 & 41.30$\pm$0.27 & 43.39$\pm$0.26 & 44.56$\pm$0.83 & +3.25 & 58.5 \\
task\_unseen & 803 / 4,464 & 43.03$\pm$0.70 & 45.79$\pm$0.68 & 46.87$\pm$1.00 & +3.84 & 59.6 \\
category\_unseen & 700 / 3,891 & 41.71$\pm$0.45 & 44.06$\pm$0.44 & 45.20$\pm$0.83 & +3.48 & 57.4 \\
\midrule
OOD union & 827 / 4,586 & 43.21$\pm$0.55 & 45.65$\pm$0.60 & 46.74$\pm$0.94 & +3.52 & -- \\
Full test & 1,543 / 8,444 & 49.12$\pm$0.64 & 51.81$\pm$0.62 & 52.85$\pm$0.94 & +3.73 & -- \\
\midrule
Validation & 137 / 690 & 57.39$\pm$1.67 & 60.43$\pm$1.74 & 61.30$\pm$2.13 & +3.91 & -- \\
\bottomrule
\end{tabular}
\caption{\textbf{Per-subsplit waterfall on \achigh{}, with the a11y-text PaLM~2S cells (closest published protocol).} Exact-match step accuracy of the base policy, after the goal-source span ranker and after the type-gated click reranker (final system); mean $\pm$ sd over 3 training seeds on identical base checkpoints (per seed: \cref{tab:seeds}). $\Delta$ is final minus base; $\Delta$ and all differences in the text use unrounded means. The OOD union is derived per seed from the full-test and IDD results; the validation split selected the reranker's variant, $\beta$ and gate. PaLM~2S: the LoRA-tuned a11y-text model of \citet[Tab.~5]{li2024effects} on the same subsplits under the original protocol. The OOD subsplits overlap (\cref{app:protocol}). \aclow{}: \cref{tab:subsplit_waterfall_low}.}
\label{tab:subsplit_waterfall}
\end{table*}

\begin{table*}[t]
\centering\footnotesize
\setlength{\tabcolsep}{4pt}

\begin{tabular}{lcccc}
\toprule
Subsplit & Base & + ranker & + reranker & PaLM 2S \\
\midrule
IDD & 82.77 & 82.68 & 83.09$\pm$1.01 & 83.2 \\
app\_unseen & 72.99 & 71.75 & 72.37$\pm$1.58 & 78.5 \\
task\_unseen & 73.66 & 72.65 & 73.15$\pm$1.46 & 77.3 \\
category\_unseen & 72.84 & 71.75 & 72.30$\pm$1.68 & 76.8 \\
\midrule
OOD union & 73.69 & 72.51 & 73.04$\pm$1.44 & -- \\
Full test & 77.84 & 77.16 & 77.64$\pm$1.23 & -- \\
\midrule
Validation & 83.91 & 83.48 & 84.16$\pm$0.89 & -- \\
\bottomrule
\end{tabular}
\caption{\textbf{Per-subsplit waterfall on \aclow{}} (same models and conventions as \cref{tab:subsplit_waterfall}; sd for the final system, 0.8--1.7 at every stage). The PaLM~2S \aclow{} setting also receives the episode goal.}
\label{tab:subsplit_waterfall_low}
\end{table*}

\begin{figure*}[t]
  \centering
  \includegraphics[width=\textwidth]{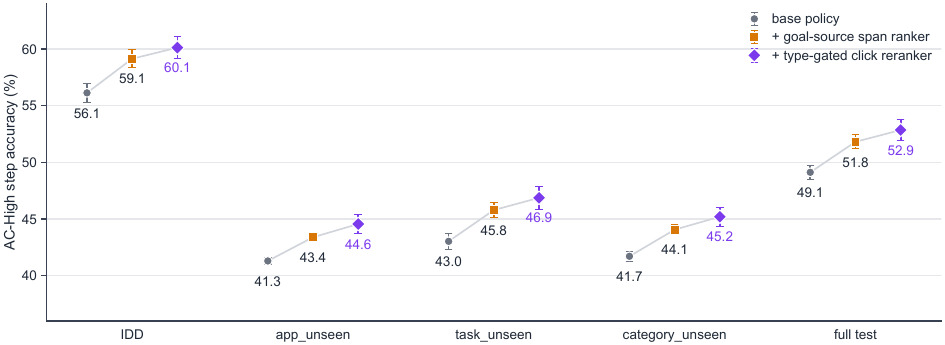}
\caption{\textbf{The text rankers' gains transfer across subsplits.} \achigh{} step accuracy (3 training seeds, mean $\pm$ sd) of the base policy, after the span ranker and after the click reranker, on IDD, the three overlapping OOD subsplits and the full test set; increments are computed on identical base checkpoints. Numbers: \cref{tab:subsplit_waterfall}.}
\label{fig:waterfall}
\end{figure*}

\Cref{tab:subsplit_waterfall,tab:subsplit_waterfall_low} and \cref{fig:waterfall} break the waterfall of \cref{tab:waterfall} down by subsplit. The \achigh{} gains transfer: on the three OOD subsplits the span ranker adds 2.08--2.76 points, and the click reranker adds 1.04--1.17 on the full test set, the OOD union and every OOD subsplit, at least its 0.98 on IDD. Neither stage changes the action-type decision of the base policy, but \aclow{} pays a small cost. The single span ranker, trained on goals, also supplies the typed argument for \aclow{} step instructions and lowers \aclow{} on every seed (by 1.18 points on the OOD union): on the full test set it raises High \texttt{input\_text} accuracy from 17.7 to 53.7 (IDD: 12.3 to 53.4) but lowers Low \texttt{input\_text} accuracy from 83.4 to 74.4. The click reranker, which raises IDD High click grounding given the type from 71.8 to 73.7, also runs in \aclow{} and recovers 0.53 of the 1.18 OOD-union points, leaving OOD-union \aclow{} 0.65 below the base policy.

On the validation split the waterfall keeps the test ordering and step sizes (+3.04 and +0.87 High). The span-ranker step is thus a held-out replication of a choice made on test IDD. The reranker step is not, because the reranker's variant, $\beta$ and gate were selected on this split; its held-out evidence is the test-set gain.

The PaLM~2S columns give the only published baseline that consumes the a11y tree (text only, no screenshot) with our click rule and subsplits; its \aclow{} setting receives the episode goal together with the step instruction \citep{li2024effects}, while ours receives the step instruction only. Averaged over the three seeds, \lwm{} matches it on IDD \aclow{} (83.09 vs 83.2) and trails by 10.7 \achigh{} points; out of distribution it trails by 12.2--13.9 High and 4.2--6.1 Low points.

\section{Mechanism analyses}
\label{app:mechanism}

\begin{table*}[t]
\centering\footnotesize
\setlength{\tabcolsep}{3pt}

\begin{tabular}{@{}>{\raggedright\arraybackslash}p{4.9cm}ccccc@{}}
\toprule
 & \multicolumn{2}{c}{IDD} & app\_unseen & task\_unseen & category\_unseen \\
\cmidrule(lr){2-3}\cmidrule(lr){4-4}\cmidrule(lr){5-5}\cmidrule(lr){6-6}
Condition & \aclow{} & \achigh{} & Low / High & Low / High & Low / High \\
\midrule
\rowcolor{oursrow}Base policy (as is) & 82.8$\pm$1.0 & 56.2$\pm$0.8 & 72.9 / 40.8 & 73.6 / 42.6 & 72.7 / 41.3 \\
\quad no predictor ($\zhat^{c}:=\zt$) & 61.0$\pm$1.7\dn{21.9} & 41.7$\pm$1.1\dn{14.6} & 53.8 / 32.6 & 53.7 / 33.1 & 52.8 / 32.2 \\
\quad within-type action shuffle & 71.9$\pm$0.8\dn{10.9} & 45.8$\pm$0.8\dn{10.4} & 62.1 / 32.3 & 63.5 / 34.5 & 61.9 / 32.7 \\
\quad all-candidate action shuffle & 56.1$\pm$0.5\dn{26.7} & 34.4$\pm$0.9\dn{21.8} & 49.9 / 26.7 & 50.3 / 27.6 & 49.3 / 26.8 \\
\quad swapped, non-fine-tuned predictor & 66.5$\pm$1.9\dn{16.3} & 45.5$\pm$0.5\dn{10.7} & -- & -- & -- \\
\quad text matcher dropped (structural head alone) & 38.4$\pm$1.7\dn{44.5} & 30.6$\pm$1.2\dn{25.7} & -- & -- & -- \\
\midrule
Frozen dynamics-pretrained predictor & 80.3$\pm$0.3\dn{2.5} & 54.9$\pm$0.1\dn{1.3} & -- & -- & -- \\
\rowcolor{oursrow}Jointly fine-tuned predictor (base) & 82.8$\pm$0.9 & 56.1$\pm$0.8 & -- & -- & -- \\
Randomly initialised predictor, joint & 82.6$\pm$0.3\dn{0.2} & 57.0$\pm$0.8\up{0.9} & -- & -- & -- \\
Identity map in place of the predictor, retrained & 62.3$\pm$0.7\dn{20.5} & 42.1$\pm$0.8\dn{14.0} & 55.0 / 31.6 & 55.5 / 32.2 & 54.3 / 31.2 \\
Matcher only (no latent pathway), retrained & 55.3$\pm$0.3\dn{27.4} & 39.9$\pm$0.7\dn{16.2} & 47.7 / 29.6 & 47.2 / 30.5 & 46.8 / 29.6 \\
No predictor, action vector into a param.-matched head (per seed) & 82.4 / 82.0 / 53.9 & 56.5 / 56.8 / 41.1 & 72.4 / 40.4$^{\S}$ & 73.2 / 41.9$^{\S}$ & 72.4 / 40.9$^{\S}$ \\
\midrule
Per-patch next-latent loss & -- & $+0.26$ (mean) & -- & -- & -- \\
Pooled next-latent MSE & -- & $+0.6$ / $-13.4$ / $+1.3$ & -- & -- & -- \\
Multi-horizon InfoNCE & -- & $+0.7$ / $-1.2$ & -- & -- & -- \\
SIGReg-style (LeWorldModel loss) pretrained predictor & -- & $-1.0$ & -- & -- & -- \\
\bottomrule
\end{tabular}
\caption{\textbf{Ablations of the latent-predictor pathway} on the full-scale base policy (exact-match step accuracy; IDD mean $\pm$ sd, OOD cells means \aclow{} / \achigh{}; subscripts: change against the as-is row (top) or the jointly fine-tuned base (middle)). Top: inference-time lesions on all steps of each split, text matcher untouched except in the last row; the as-is row is an independent evaluation of the same policies (on IDD within 0.2 of \cref{tab:seeds}). Middle: training-time variants (base-policy OOD: the as-is row); the two retrained rows use the identical full-scale recipe (OOD on all steps); the parameter-matched row lists its three training runs on IDD, and $^{\S}$marks the mean of the two that train stably (the third collapses on clicks: 45.5--47.7 Low / 28.8--31.4 High out of distribution). Bottom: predictive objectives at 4K-episode scale (auxiliary losses with $\lambda{=}0.1$; multi-horizon InfoNCE 0.2), \achigh{} change against the matched 4K reference, paired per seed unless a mean is stated. Per-seed IDD values: \cref{tab:seeds_mech}; concat-MLP head (pilot scale): \cref{tab:negatives}.}
\label{tab:lesion}
\end{table*}

\Cref{tab:lesion} collects the evidence on the latent-predictor pathway summarised in \cref{sec:analysis_pathway}; the text below adds only what the table does not show.

\paragraph{Inference-time lesions.}
The lesions are applied at evaluation time to the trained full-scale base policies (text rankers off), so they measure how much the trained scorer relies on the pathway; the retrained controls below test whether it is needed. Removing the predictor ($\zhat^{c}:=\zt$) hurts mainly the global actions: of its 21.9 \aclow{} / 14.6 \achigh{} IDD points, 19.0 / 14.1 fall on scroll, \texttt{open\_app} and \texttt{navigate\_back} (19.6 / 15.6 with \texttt{wait}). These share the full-screen ROI, so without the predictor their action vectors no longer reach the score and the three types are never chosen correctly, while click accuracy barely moves (83.9 to 83.2 Low, 58.4 to 60.6 High). The within-type shuffle, which permutes the four scroll directions, costs 8.6 / 6.3 points on scroll steps and 2.3 / 4.2 on clicks. Without the text matcher the structural head's click grounding is 30.9 Low / 31.8 High against 89.8 / 71.8 for the full scorer: the matcher carries most of the click grounding, while the head alone still grounds about a third of the click steps whose type it gets right. The reliance holds on every OOD subsplit; for the main-text model, removing the predictor costs 21.7--23.1 Low / 7.9--9.7 High points there, the within-type shuffle 10.5--11.5 / 8.5--9.0 and the all-candidate shuffle 24.4--25.0 / 15.4--16.7 (means: \cref{tab:lesion}).

\paragraph{Training-time variants.}
Freezing the dynamics-pretrained predictor costs 2.52 \aclow{} / 1.25 \achigh{} IDD points on average (3.78 / 1.89 for seed 3; \cref{tab:seeds_mech}). A randomly initialised predictor trained jointly with the identical full-scale recipe matches the dynamics-pretrained one: its High is higher for every seed (+0.50 to +1.17) and its Low within 1.01 points. At pilot scale, the identity map gives 31.73 Low vs 40.59 for the ROI-conditioned predictor (structural head, no text matcher; \cref{fig:mechanism}b), and training on shuffled dynamics with a frozen predictor gives 47.33 / 28.46 vs 62.70 / 38.96.

\paragraph{Retrained controls.}
Retrained from scratch at full scale with the identical recipe, the identity map in place of the predictor again never chooses scroll, \texttt{open\_app} or \texttt{navigate\_back} correctly ($-11.3$, $-5.1$ and $-2.6$ Low points), while clicks change by $-1.8$ Low / $+0.2$ High points. The matcher-only policy scores 0--12\% on scroll, \texttt{open\_app}, \texttt{navigate\_back}, \texttt{wait} and \texttt{input\_text}, and its click grounding is 88.4 / 68.2 against the base policy's 89.8 / 71.8. A no-predictor control tests whether the predictor is needed at all once the action reaches the score: its structural head receives $[\pool_{\roi(c)}(\zt);\,\mathrm{mean}(\zt);\,a_c]$ through an MLP matched in parameters to the predictor and the original projection (41.09M against 39.91M + 1.18M), with matcher, history, type bias and recipe unchanged. It matches the base policy in two of three seeds (\cref{tab:seeds_mech}; within 1.4 points on app- and category-unseen) and collapses on clicks in the third (click grounding 48.3 Low against 89.6 for its base seed). Given the action vector directly, a parameter-matched head on the same JEPA latents thus reaches the base policy's accuracy when its training succeeds, whereas the predictor trains stably in all three seeds; either way the action is scored on the JEPA latent, consistent with the predictor's role as an action- and ROI-conditioned transform of it (\cref{sec:analysis_pathway}).

\paragraph{Predictive objectives.}
The bottom block of \cref{tab:lesion} gives the predictive auxiliary objectives as \achigh{} changes against the matched 4K reference; none improves on it (\cref{app:negatives}).

\begin{table*}[tbp]
\centering\small
\setlength{\tabcolsep}{4pt}

\begin{tabular}{lccc}
\toprule
 & IDD & app\_unseen & task\_unseen \\
\midrule
Probe steps / episodes & 1,579 / 300 & 116 / 24 & 1,671 / 300 \\
cos(prediction, true next latent) & 0.467 & 0.467 & 0.460 \\
cos(copy $\zt$, true next latent) & 0.788 & 0.773 & 0.786 \\
cos(predicted change, true change) & 0.006 & 0.003 & 0.004 \\
Right direction (\%) & 52.6 & 47.4 & 51.6 \\
\bottomrule
\end{tabular}
\caption{\textbf{Next-latent probes of the deployed (jointly fine-tuned) predictor.} ``Copy'' keeps the current latent $\zt$ as the prediction; direction is the share of steps on which the predicted latent change points the same way as the true change (chance 50\%).}
\label{tab:e1probe}
\end{table*}

\paragraph{Forward-model probes.}
\Cref{tab:e1probe} shows that the deployed predictor is a poor forward model: its output is further from the true next latent than a copy of the current latent, and its predicted change has near-zero cosine to the true change and points the right way at chance rate. Feeding a shuffled action leaves the predicted latent at cosine 0.86 to the true-action prediction (IDD). The randomly initialised predictor of \cref{tab:lesion} (150 IDD episodes, 782 steps) is no more action-sensitive: its predicted change has cosine 0.002 to the true change, its ground-truth AUC equals the deployed predictor's (0.5426 vs 0.5425; chance 0.5), and a shuffled action leaves its output at cosine 0.887 (0.954 for a within-type shuffle, probed for this predictor only). Free-running $K$-step rollouts of the dynamics-pretrained predictor that initialises the deployed one (400 episodes; 1,500 / 1,500 / 1,349 queries at $K{=}1$ / 2 / 3), scored by cosine to the true latent $K$ steps ahead, never beat persistence (keeping $\zt$): 0.745 against 0.793 at $K{=}1$ (0.733 with shuffled actions), 0.510 against 0.749 at $K{=}2$ and 0.322 against 0.741 at $K{=}3$. The rollout beats persistence on 39 / 10 / 2\% of steps and a shuffled-action rollout on only 43 / 48 / 49\%; only within-episode retrieval of the true frame at $K{=}1$ slightly favours it (R@5 79.7 vs 77.2 for persistence). Its one-step cosine (0.745) exceeds the deployed predictor's 0.467, consistent with joint fine-tuning on the policy loss moving the predictor away from next-latent prediction; both are below copying (0.793 / 0.788).

\section{Design alternatives not adopted}
\label{app:negatives}

\begin{table*}[t]
\centering\footnotesize
\setlength{\tabcolsep}{3pt}

\resizebox{\textwidth}{!}{%
\begin{tabular}{p{4.4cm}p{2.3cm}ccp{4.0cm}}
\toprule
Variant & Setting & Reference & Variant & Reading \\
\midrule
\multicolumn{5}{l}{\emph{Scoring head}} \\
{\raggedright Concat-MLP head over $[\zt,\zhat^{c},c]$ replacing the structural head (ROI-conditioned predictor)\par} & pilot, 2 ep & 40.59 / 19.62 & 34.58 / 26.88 & {\raggedright $-6.0$ Low, $+7.3$ High; its scores do not depend on the predictor\par} \\
\midrule
\multicolumn{5}{l}{\emph{Encoder and resolution}} \\
{\raggedright DINOv2-L at 518\,px as main features\par} & 4K, 4 ep & 77.32 / 47.12 & 59.88 / 37.35 & {\raggedright encoder and resolution change together\par} \\
\bottomrule
\end{tabular}}
\caption{\textbf{Design alternatives not adopted} (\aclow{} / \achigh{} exact-match step accuracy). Each variant is compared with a reference model that differs from it only in the stated factor and shares its scale, epochs and evaluation split. Setting: pilot = early pilot configuration (no text matcher); 4K = 4,000-episode training subset; ep = policy training epochs; evaluation on IDD (3,858 steps). Predictive objectives for the predictor: \cref{tab:lesion}.}
\label{tab:negatives}
\end{table*}

\Cref{tab:negatives} lists the design alternatives not adopted; the paragraphs below add what it does not show.

\paragraph{Scoring head.}
The concat-MLP head trades \aclow{} for \achigh{} accuracy: against the structural head it loses \aclow{} and gains \achigh{} points with the ROI-conditioned predictor, and likewise loses 7.3 Low and gains 8.8 High with an $(x,y)$-conditioned predictor (40.51 / 18.43 vs 33.18 / 27.19). Its \aclow{} accuracy is the same with either predictor or none (\cref{fig:mechanism}b), so whatever it gains does not come through the predictor, whereas the structural head drops from 40.59 to 31.73 when the predictor is replaced by the identity map. We keep the structural head because it forces the policy through the candidate-conditioned latent; the higher pilot-scale High of the concat head shows that this constraint has a cost at that scale. Conditioning the structural head on the previous action and frame change (implicit action history) was adopted: it gains 8.0 High at pilot scale without the text matcher (23.35 to 31.39, at a cost of 4.4 Low) and 7.1 High with it (31.91 to 38.96, $-1.6$ Low), and the final system uses it together with the textual action history in the matcher query.

\paragraph{Encoder and resolution.}
Swapping in DINOv2-L features \citep{oquab2024dinov2learningrobustvisual} changes the encoder and the resolution (518\,px) together, so this comparison does not isolate the effect of resolution.

\paragraph{Predictor objectives.}
Predictive auxiliary losses on the predictor do not help (\cref{tab:lesion}, bottom). The leakage-free pooled next-latent MSE has a low mean because one of its three runs collapses; the other two are within 1.3 points of their paired controls. Our SIGReg-style regulariser simplifies SIGReg \citep{balestriero2025lejepaprovablescalableselfsupervised}: where LeJEPA uses the Epps--Pulley statistic, it penalises the mean and variance of 256 random projections of the pooled predicted latents, and it acts on frozen encoder features, whereas LeWorldModel \citep{maes2026leworldmodel} trains end to end.

\section{Other-protocol baselines}
\label{app:baselines}

\begin{table*}[t]
\centering\footnotesize
\setlength{\tabcolsep}{3pt}

\begin{tabular}{@{}p{4.0cm}lp{1.4cm}p{2.0cm}p{1.6cm}cc@{}}
\toprule
System (source) & Params & Input & Judge & Split & \aclow{} & \achigh{} \\
\midrule
HiconAgent-3B \citep[Tab.~1]{DBLP:conf/cvpr/ZhouCXLZWYTS26} & 3B & img & OS-Atlas & unspec. (``OOD setting'') & -- & 52.40 \\
UI-TARS-2B \citep[Tab.~8]{qin2025ui} & 2B & img & OS-Atlas & 7,708 (inferred) & 89.3 & 68.9 \\
LAMO-3B \citep[Tab.~2]{Wang2026TowardsSL} & 3B & img & own (type + value match) & unspec. & 92.1 & 65.5 \\
SecAgent-3B \citep[Tab.~3]{xie2026secagentefficientmobilegui} & 3B & img & unspec. & unspec. & -- & 69.5 \\
GUI-Libra-3B \citep[Tab.~3]{yang2026guilibratrainingnativegui} & 3B & img & own & AC-v2 (398) & 85.9 & 57.3 \\
Qwen2.5-VL-3B, re-run by GUI-Libra \citep[Tab.~3]{yang2026guilibratrainingnativegui} & 3B & img & own & AC-v2 (398) & 71.1 & 36.4 \\
UI-R1-3B \citep[Tab.~3]{lu2026uir1} & 3B & img & SR, clicks within 14\% of screen & unspec. & 66.4 & -- \\
LiMAC (AcT + Florence-2) \citep[Tab.~1]{christianos2025lightweight} & $\approx$1.34B & img + UI tree & own (text Jaccard $\ge$0.5) & unspec. & -- & 63.1$^{\dagger}$ \\
Florence-2 alone (LiMAC baseline) \citep[Tab.~1]{christianos2025lightweight} & 0.82B & img + UI-tree boxes & own (text Jaccard $\ge$0.5) & unspec. & -- & 57.0$^{\dagger}$ \\
Qwen2-VL-2B, LoRA (LiMAC baseline) \citep[Tab.~1]{christianos2025lightweight} & 2B & img + UI-tree boxes & own (text Jaccard $\ge$0.5) & unspec. & -- & 52.2$^{\dagger}$ \\
\bottomrule
\end{tabular}
\caption{\textbf{Small-model \ac{} results not in \cref{tab:main}} (larger and zero-shot models under other protocols: \cref{tab:otherprotocol2}). Every cell is copied from the cited table. Apart from LiMAC and its two baselines, discussed in the text, none is set against our numbers because input, judge or split differ (ours for reference: 77.64 / 52.85 Low / High on the full test and 83.09 / 60.12 on IDD as 3-seed means, 78.64 / 53.75 and 84.06 / 60.99 for seed 3, under the exact-match judge with a11y input). Judge: \emph{relaxed} = original protocol; \emph{OS-Atlas} = a click within 14\% of the screen width counts and typed text at F1$>$0.5 \citep[App.~D]{DBLP:conf/iclr/0003WXWSJC0CL025}; \emph{own} = the paper's own definition. Split: \emph{7,708} = OS-Atlas test samples; \emph{r500} / \emph{r800} = a random 500- / 800-step test subset; \emph{r500-IDD} = IDD part of Random-500; \emph{unspec.} = not stated in the source; \emph{AC-v2 (398)} = the 398-example cleaned subset of GUI-Libra (its Pass@1 columns). $^{\dagger}$Goal-conditioned per-step action accuracy (``overall accuracy''), the analogue of \achigh{}. Parameter counts are nominal.}
\label{tab:otherprotocol}
\end{table*}

\begin{table*}[t]
\centering\footnotesize
\setlength{\tabcolsep}{3pt}

\begin{tabular}{@{}p{4.0cm}lp{1.4cm}p{2.0cm}p{1.6cm}cc@{}}
\toprule
System (source) & Params & Input & Judge & Split & \aclow{} & \achigh{} \\
\midrule
GPT-4o, zero-shot \citep[Tab.~5]{DBLP:conf/iclr/0003WXWSJC0CL025} & n/a & img & OS-Atlas & 7,708 & 28.4 & 21.2 \\
OS-Atlas-7B, zero-shot \citep[Tab.~5]{DBLP:conf/iclr/0003WXWSJC0CL025} & 7B & img & OS-Atlas & 7,708 & 50.9 & 29.8 \\
M3A (GPT-4-Turbo, text-only), zero-shot \citep[Tab.~4]{li2024effects} & n/a & a11y & relaxed & r500-IDD & 55.0 & 42.1 \\
Aguvis-7B, SFT \citep[Tab.~3]{DBLP:conf/icml/XuWWLXSS0X25} & 7B & img & relaxed (inferred) & r500 & 80.5 & 61.5 \\
Aguvis-72B, SFT \citep[Tab.~3]{DBLP:conf/icml/XuWWLXSS0X25} & 72B & img & relaxed (inferred) & r500 & 84.4 & 66.4 \\
OS-Atlas-7B, SFT \citep[Tab.~5]{DBLP:conf/iclr/0003WXWSJC0CL025} & 7B & img & OS-Atlas & 7,708 & 85.22 & 71.17 \\
UI-TARS-7B, SFT \citep[Tab.~8]{qin2025ui} & 7B & img & OS-Atlas & 7,708 (inferred) & 90.8 & 72.5 \\
MAI-UI-8B / 32B \citep[Tab.~7]{Zhou2025MAIUITR} & 8B / 32B & img & unspec. & unspec. & -- & 69.1 / 75.5 \\
ScaleTrack-7B \citep[Tab.~3]{huang2025scaletrackscalingbacktrackingautomated} & 7B & img & unspec. & r800 (``out-of-domain'') & 86.6 & 77.9 \\
\bottomrule
\end{tabular}
\caption{\textbf{Zero-shot and larger models under other protocols, or with \achigh{} only} (conventions and judge / split definitions as in \cref{tab:otherprotocol}). Every cell is copied from the cited table; none is set against our numbers. MAI-UI reports exact match on an unstated split and no \aclow{}.}
\label{tab:otherprotocol2}
\end{table*}

\Cref{tab:otherprotocol,tab:otherprotocol2} list the \ac{} results of small models and of larger or zero-shot systems reported under protocols that \cref{tab:main} does not include; their judge, split and input columns show why none except LiMAC is set against our numbers. None of the larger models is evaluated on the full test set we use.

LiMAC \citep{christianos2025lightweight} is the closest small agent that also consumes the UI tree: an action transformer (AcT, $\approx$520M) over UI-element embeddings combined with Florence-2 \citep{Xiao_2024_CVPR} (820M as deployed by LiMAC), which receives the screenshot annotated with numbered element boxes and the goal. Its overall accuracy is a goal-conditioned per-step action accuracy, the analogue of our \achigh{} setting, under its own judge (a predicted element whose box lies within the target element is correct, typed text is accepted at a Jaccard index of at least 0.5) on ``the test set'', with no sub-split or size given; its confusion matrix sums to 3,708 steps, matching none of the official sub-splits. LiMAC and Florence-2 alone are 10.3 and 4.2 points above our 3-seed mean full-test High (9.3 and 3.2 above seed 3) with 2.7$\times$ and 1.6$\times$ our parameters, and its fine-tuned Qwen2-VL-2B baseline is 0.65 below our mean with 4.0$\times$ our parameters; LiMAC's more lenient text rule and unstated split keep the comparison approximate. LiMAC reports 0.34\,s per step (Florence-2 alone: 0.50\,s) on unspecified hardware, against our 51.5--74\,ms per real step on one H20 (\cref{app:latency}).

With the three-seed mean full-test \aclow{} (77.64; seed 3: 78.64), \lwm{} is level with Qwen3-VL-8B (77.7) and MIRAGE-4B (77.6) of \cref{tab:main}.

\section{Encoder probe}
\label{app:encoder}

\begin{figure*}[t]
  \centering
  \includegraphics[width=\textwidth]{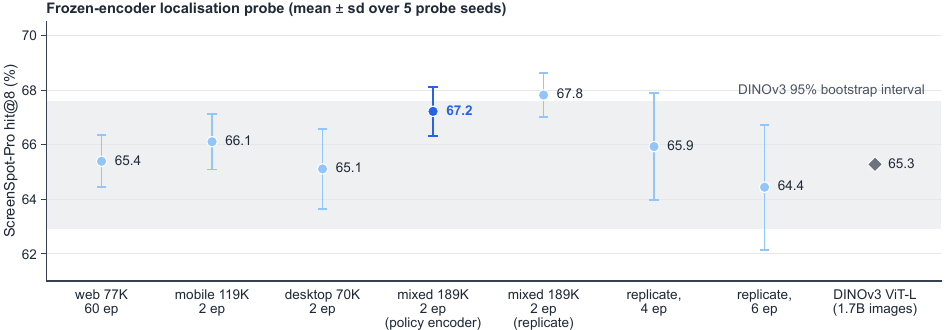}
\caption{\textbf{Lenient frozen-encoder localisation probe across pretraining recipes.} ScreenSpot-Pro hit@8 of a soft-argmax localisation probe on frozen patch tokens (our encoders: mean $\pm$ sd over 5 probe seeds); hit@8 accepts errors up to 57\% of the image side. Points, left to right: a web anchor ($\approx$77K web screenshots, 60 epochs); mobile-only (119K GUI-Odyssey screenshots, 2 epochs); desktop-only (70K VideoCUA screenshots, 2 epochs); the policy encoder (blue; the mixed 189K mobile and desktop recipe, 2 epochs); a replicate pretraining of the same recipe after 2 epochs and, continued, after 4 and 6 epochs; DINOv3-ViT-L with its 95\% bootstrap interval over probe examples (band). The recipes also differ in data size and training length, and the probe measures coarse localisation only (see text).}
\label{fig:encoder}
\end{figure*}

\paragraph{Probe.}
\Cref{fig:encoder} compares pretraining recipes for the frozen encoder by a localisation probe on its patch tokens: a linear scorer per patch, a softmax over the $14\times14$ grid and a soft-argmax give a point that is regressed (MSE) onto the centre of the target box; the encoder is fixed and only the probe is trained. hit@$k$ is the fraction of held-out examples whose predicted point lies within $k/14$ of the image side (normalised Euclidean distance) of the target centre, so hit@8 accepts errors of up to 128\,px on the 224-px input. The probe is lenient: on the 1,581 held-out ScreenSpot-Pro \citep{Li2025ScreenSpotProGG} examples the policy encoder's hit@8 of 67.22 $\pm$ 0.91 comes with a hit@3 of 10.36 $\pm$ 0.36, a hit@1 of 0.92 $\pm$ 0.36 and a mean error of 0.509 of the image side (114\,px); on 255 ScreenSpot-v2 \citep{DBLP:conf/iclr/0003WXWSJC0CL025} examples it reaches 90.59 $\pm$ 1.00 hit@8 with a mean error of 0.378 of the side (85\,px). For calibration, under the same probe a randomly initialised ViT-L reaches 87.8 hit@8 on the same ScreenSpot-v2 split and 58.8 [56.4, 61.2] on ScreenSpot-Pro, and an ImageNet-initialised ViT-L without GUI pretraining 58.1. The probe thus measures coarse localisation.

\paragraph{Recipes.}
Under this probe the mixed recipe, I-JEPA-style pretraining \citep{assran2023self} of an ImageNet-initialised ViT-L on 119K GUI-Odyssey \citep{Lu_2025_ICCV} mobile and 70K VideoCUA \citep{jian2026cuasuitemassivehumanannotatedvideo} desktop frames for 2 epochs, scores at least as high as every other recipe in \cref{fig:encoder}: either domain alone, a web anchor trained for 60 epochs on the 76,838 viewport and full-page screenshots of the class-balanced 7K split (Web-7k-Resampled) of WebUI \citep{DBLP:conf/chi/WuWSPNB23}, and DINOv3-ViT-L \citep{simeoni2026dinov3}, trained on 1.7B images. DINOv3's 95\% bootstrap interval over probe examples, [62.9, 67.6], contains the policy encoder's 67.22 (95\% $t$-interval over probe seeds [66.1, 68.3]): under this probe the policy encoder is at DINOv3's level, and since the two intervals cover different sources of variance (probe seeds against probe examples), they do not order the encoders. The replicate uses the same data, batch size and learning-rate schedule and continues to 6 epochs: after 2 epochs (JEPA loss 0.668 against 0.677 for the policy encoder) it gives 67.82, and its 4- and 6-epoch scores show that training longer does not help. The 0.6-point gap between the two 2-epoch encoders is a rough measure of pretraining variance, which the per-recipe sds, computed over probe seeds, do not capture. The recipes also differ in data size (189K / 119K / 70K / $\approx$77K screenshots) and training length (2 to 60 epochs), so the comparison does not identify which ingredient matters. Two further recipes do not lift the probe mean: initialising JEPA pretraining from DINOv3 lowers it, and action-conditioned pretraining in the style of V-JEPA~2-AC \citep{Assran2025VJEPA2S} on VideoCUA stays at the level of the web anchor.

The probe measures frozen localisation features: the recipe is a 0.76 GPU-hour route to frozen GUI features that localise as well as the alternatives. In the policy, the 4K-scale swap to DINOv2-L at 518\,px (\cref{tab:negatives}) changes the encoder and the resolution together.

\subsection{Pretraining--test near-duplicates}
\label{app:overlap}
We hashed the 9,987 screenshots of the 1,543 \ac{} test episodes (pHash and dHash, 64 bits) against the 189,262 pretraining screenshots (118,791 GUI-Odyssey, 70,471 VideoCUA). 58 test frames (0.58\%) in 43 episodes have a verified near-duplicate (pHash distance $\le 8$ and normalised cross-correlation $\ge 0.9$), all from GUI-Odyssey and showing generic system screens. For calibration, the same pHash threshold matches 35.0\% of test frames to the frames of 4,000 \ac{} training episodes (different episodes of the same benchmark) and 9.5\% to the pretraining set; random pairs have a median distance of 32. Removing the 43 episodes changes final High and Low by $-0.2$ (High 59.90 vs 60.12 on IDD, 52.64 vs 52.85 on the full test). Removing every episode with any pHash match within distance 8 (549 episodes), a criterion that also flags generic layouts, gives 58.74 IDD and 51.14 full-test High. At the app level, 29 app names match exactly and 26\% of test episodes touch a package that also appears in GUI-Odyssey, with different devices and tasks and no shared screenshots. Pretraining--test leakage is thus negligible for the reported numbers.

\section{Implementation details and hyperparameters}
\label{app:implementation}

\begin{table*}[tbp]
\centering\footnotesize
\setlength{\tabcolsep}{4pt}

\begin{tabular}{llp{7.4cm}}
\toprule
Component & Params & Setting \\
\midrule
\multicolumn{3}{l}{\emph{Encoder pretraining (frozen afterwards)}} \\
JEPA encoder, ViT-L/16 & 303.3M & \texttt{vit\_large\_patch16\_224}, 224\,px, $14\times14=196$ patch tokens; ImageNet initialisation; JEPA objective after I-JEPA with random-patch masking at ratio 0.5 (the student sees the screenshot with the masked patches zeroed and regresses the EMA target's tokens at those positions, smooth-L1); 189,262 unlabelled screenshots (118,791 GUI-Odyssey mobile, 70,471 VideoCUA desktop); 2 epochs at batch 96, 3,942 optimiser steps (the policy uses the epoch-2 checkpoint); EMA target decay 0.996; peak learning rate $5\times10^{-5}$ with warm-up; JEPA loss 1.28 $\to$ 0.68; target-token std 1.8--2.2 (min 1.82, final 2.17; no collapse); 2,739\,s $\approx$ 0.76 GPU-hour on one H20 \\
\midrule
\multicolumn{3}{l}{\emph{Policy (trained end to end with the listwise loss)}} \\
Action code $a_c$ & -- & 16-d: action-type one-hot plus normalised coordinates or scroll direction \\
Residual latent predictor & 39.9M & ROI-pooled encoder feature $\mathbf{u}_c$ (1,024-d) $\oplus$ $a_c$ $\to$ MLP $\to$ added to a learned query token; 3 transformer layers over the 196 patch tokens and the query, 2-D sine--cosine positional embeddings; residual output $\zhat^{c}=\zt+\Delta(\zt,a_c,\mathbf{u}_c)$; initialised from a dynamics predictor trained on \ac{} transitions, then fine-tuned with the policy loss \\
Structural head & 2.1M & cosine, with a learned scale, between an MLP of $[\pool_{\roi(c)}(\zhat^{c});\,\mathrm{mean}(\zhat^{c})]$ and an MLP of the conditioning vector, plus a learned per-type bias; its only candidate-dependent visual input is $\zhat^{c}$ \\
Conditioning vector & 63.2M & frozen CLIP text encoder: embedding of the goal (High, once per episode) or of the step instruction (Low, once per step), joined with the previous action code and the mean latent difference between the previous and the current frame \\
Text matcher & 33.4M & MiniLM-L12 cross-encoder over the instruction (with the textual action history appended) and the candidate text; its score is added outside the structural head \\
\midrule
\multicolumn{3}{l}{\emph{Post-hoc modules}} \\
Goal-source span ranker & 22.7M & MiniLM-L6 cross-encoder over spans of the goal, on which it is also trained; supplies the \texttt{input\_text} argument in both settings \\
Type-gated click reranker & 33.5M & MiniLM-L12 listwise reranker over the click candidates; $\mathrm{final}=\mathrm{base}+\beta\,\tilde r$ where $\tilde r$ is the $z$-scored reranker logit, applied only when the base argmax is a click and the reranker confidence exceeds the gate, in both settings; variant, $\beta{=}1$ and gate $=1$ selected on the validation split \\
\midrule
\multicolumn{3}{l}{\emph{Optimisation of the base policy}} \\
Data & -- & \ac{} training split as cached, 13,594 of 13,604 episodes / 74,714 of 74,722 steps (\cref{app:protocol}); cached frozen features; candidates capped at 80 per step \\
Loss & -- & listwise cross-entropy over the enumerated candidates with the ground-truth candidate as positive; weight 1.5 on non-click ground-truth types, 1 on clicks \\
Optimiser & -- & AdamW; learning rate $10^{-4}$ for predictor and heads, $2\times10^{-5}$ for MiniLM; weight decay 0.01; no dropout; 6 epochs; 3 training seeds; one model for Low and High \\
\midrule
Total & 0.50B & 498.0M: ViT-L 303.3M, CLIP text 63.2M, predictor 39.9M, head 2.1M, matcher 33.4M, click reranker 33.5M, span ranker 22.7M; 366.5M frozen (ViT-L and CLIP text) and 131.6M trained \\
\bottomrule
\end{tabular}
\caption{\textbf{Components and hyperparameters of \lwm{}.} Parameter counts are measured. The encoder and the CLIP text encoder are frozen and their outputs cached before policy training; the span ranker and the click reranker are trained after the base policy with the base policy fixed.}
\label{tab:hparams}
\end{table*}

\Cref{tab:hparams} lists every component with its measured parameter count and the settings of the reported models; the paragraphs below add only what the table does not state.

\paragraph{Encoder.}
The encoder is the ViT-L/16 of \citet{DBLP:conf/iclr/DosovitskiyB0WZ21} from the timm library. Where I-JEPA \citep{assran2023self} uses multi-block masks, our student sees the screenshot with a random half of its patches zeroed and regresses the tokens that the EMA target computes on the full image at the masked positions. The encoder is frozen and its 196 patch tokens per frame are cached; the policy never updates it.

\paragraph{Predictor, head and matcher.}
All candidates of a step form one predictor batch. The residual form was chosen in IDD pilots without the text matcher: a predictor that re-synthesises the whole latent erases local detail, lowering \aclow{} click grounding to 34.8 against 46.8 for the identity map, and the residual form raises pilot \aclow{} from 40.59 to 43.93. At pilot scale (IDD) the structural head's constraint, that its only candidate-dependent visual input is the candidate-conditioned latent, decides whether the pathway is used: a concat-MLP head over $[\zt,\zhat^{c},\mathrm{cand}]$ ignores the predictor (within 1.4 \aclow{} points), while the structural head gains 8.8 points (\cref{fig:mechanism}b; \cref{app:negatives}). The conditioning vector uses the frozen CLIP text embedding \citep{DBLP:conf/icml/RadfordKHRGASAM21}. The MiniLM-L12 text matcher \citep{DBLP:conf/nips/WangW0B0020,DBLP:conf/emnlp/ReimersG19} carries most of the click grounding: in the pilot, grounding rose from 37.4 to 70.2 when it was added. The \texttt{input\_text} argument is copied as a goal span, which can be correct on at most 81.6\% of IDD \texttt{input\_text} steps.

\paragraph{Training.}
Training skips the 1.6\% of training click steps whose ground-truth element is not enumerated. Neither text ranker changes the action-type decision of the base policy, but both act in \aclow{} as well as \achigh{} (\cref{app:waterfall}).

\begin{table*}[tbp]
\centering\small
\setlength{\tabcolsep}{3pt}

\begin{tabular}{lcccc}
\toprule
 & Qwen2.5-VL-7B & Qwen2.5-VL-7B & \lwm{} & \lwm{} \\
 & zero-shot & LoRA SFT & (ours, seed 3) & (ours, 3 seeds) \\
\midrule
Parameters & 8.29B & 8.29B & 0.50B & 0.50B \\
ms/step (H20, batch 1, 100 real steps) & 505 (HF) & 207 (vLLM) / 507 (HF) & \multicolumn{2}{c}{74 (fp32) / 51.5 (bf16)} \\
\midrule
Overall step acc. & 43.2 & 70.9 & 61.0 & 60.1$\pm$1.0 \\
\quad share of SFT (High / params) & -- & -- & 86.1\% / 6.0\% & 84.8\% / 6.0\% \\
Type accuracy & 72.1 & 87.0 & 80.6 & 78.8$\pm$3.4 \\
Click type recall & 93.5 & 91.6 & 87.1 & 83.1$\pm$7.7 \\
Click grounding (all) & 54.3 & 73.4 & 64.1 & 61.1$\pm$4.1 \\
Click grounding $\mid$ type & 58.0 & 80.1 & 73.5 & 73.7$\pm$2.3 \\
\midrule
click ($n{=}2419$) & -- & 72.0 & 62.8 & 60.0 \\
scroll (505) & -- & 71.7 & 59.8 & 62.2 \\
input\_text (282) & -- & 67.0 & 52.5 & 53.4 \\
wait (273) & -- & 57.9 & 59.3 & 62.4 \\
open\_app (253) & -- & 81.0 & 68.0 & 68.0 \\
navigate\_back (117) & -- & 64.1 & 38.5 & 49.9 \\
\bottomrule
\end{tabular}
\caption{\textbf{Same-protocol comparison with a 7B VLM agent on \achigh{} (IDD).} All systems see the same screenshots, enumerated candidates and ground-truth history and are scored with the same step-accuracy rules. The click rows use the 2,379 click and long-press steps whose ground-truth element is enumerated: \emph{type recall} is the share predicted as a click, \emph{grounding (all)} the share answered correctly with type errors counted as misses, and \emph{grounding $\mid$ type} the share correct among the steps predicted as a click. ms/step: median over the 100 real IDD steps timed for all systems (\cref{sec:efficiency}). Ours: seed 3, reported in the main text, and the mean $\pm$ sd over three seeds.}
\label{tab:sameprotocol}
\end{table*}

\paragraph{Same-protocol 7B reference.}
Qwen2.5-VL-7B-Instruct \citep{bai2025qwen25vltechnicalreport} (8.29B parameters, measured) is fine-tuned with LoRA \citep{DBLP:conf/iclr/HuSWALWWC22} on 4,000 of the 13,594 cached training episodes for one epoch, with one adapter per setting (3.5 / 4.4 GPU-h for High / Low on one H20); the \aclow{} adapter's prompt adds the step instruction to the goal and shares every other setting, whereas \lwm{} serves both settings with one model. It is prompted with the same screenshot (377 image tokens; at most 384 visual tokens), the same enumerated candidate list and the same ground-truth history as \lwm{}, decodes a JSON action greedily, and is scored by the byte-identical judge on the same 716 IDD episodes. Beyond \cref{tab:sameprotocol}, the High adapter reaches a click top-5 recall of 95.3, and the \aclow{} adapter reaches 91.01 Low with type accuracy 97.4 (\lwm{}: 84.06, 92.4\% of it). We report click grounding in two forms because the conditional form alone rewards predicting fewer clicks: our seed with the lowest type accuracy (74.9) has the highest conditional grounding (76.1). The 7B agent leads on every action type of \cref{tab:sameprotocol} except \texttt{wait}; \cref{tab:gap} converts the per-type differences into contributions to the overall gap. The fine-tuned number is close to published 7B SFT results on \ac{} (69.7--71.2 High for Qwen2-VL-7B and OS-Atlas-7B fine-tuned on \ac{}, \citealp[Tab.~5]{DBLP:conf/iclr/0003WXWSJC0CL025}), which use the OS-Atlas judge on 7,708 test samples; under our judge and split it serves as the same-protocol reference.

\begin{table}[tbp]
\centering\footnotesize
\setlength{\tabcolsep}{3.2pt}
\begin{tabular}{@{}lrccccc@{}}
\toprule
 & & \multicolumn{2}{c}{Step acc.} & \multicolumn{2}{c}{Type acc.} & \\
\cmidrule(lr){3-4}\cmidrule(lr){5-6}
Action type & $n$ & 7B & ours & 7B & ours & Gap \\
\midrule
\rowcolor{oursrow}All & 3,858 & 70.9 & 61.0 & 87.0 & 80.6 & 9.9 \\
click & 2,419 & 72.0 & 62.8 & 91.3 & 86.3 & 5.8 \\
scroll & 505 & 71.7 & 59.8 & 83.8 & 64.4 & 1.6 \\
\texttt{input\_text} & 282 & 67.0 & 52.5 & 92.2 & 87.6 & 1.1 \\
\texttt{open\_app} & 253 & 81.0 & 68.0 & 87.7 & 92.9 & 0.9 \\
\texttt{navigate\_back} & 117 & 64.1 & 38.5 & 64.1 & 38.5 & 0.8 \\
\texttt{wait} & 273 & 57.9 & 59.3 & 57.9 & 59.3 & $-$0.1 \\
\texttt{long\_press} & 9 & 44.4 & 66.7 & 100 & 88.9 & $-$0.1 \\
\bottomrule
\end{tabular}
\caption{\textbf{Where the same-protocol 7B agent leads} (\achigh{}, IDD; same inputs and judge). Step and type accuracy (\%) of the model reported in the main text by ground-truth action type; Gap: points of the overall 9.9-point gap each type contributes ($n\times$ step-accuracy difference $/$ 3,858).}
\label{tab:gap}
\end{table}

\section{Latency measurement details}
\label{app:latency}

\begin{figure}[tbp]
  \centering
  \includegraphics[width=\columnwidth]{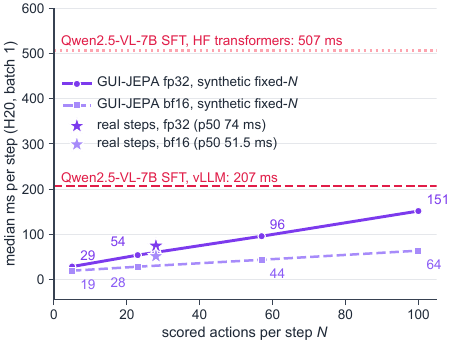}
\caption{\textbf{Same-hardware per-step latency} (median, one H20, batch 1). Stars: the deployed \lwm{} pipeline on the 100 real \achigh{} IDD steps timed for both systems, placed at their median of 28 scored actions (18 a11y candidates plus global actions). Lines: the synthetic fixed-$N$ micro-benchmark with random inputs and $N$ scored actions (\cref{tab:realstep}), a scaling illustration only. Horizontal lines: the same-protocol Qwen2.5-VL-7B SFT agent (merged LoRA) on the same steps with vLLM and HF transformers. Measurement boundaries: see the common protocol in the text.}
\label{fig:latency_sweep}
\end{figure}

\begin{table*}[tbp]
\centering\footnotesize
\setlength{\tabcolsep}{3pt}

\resizebox{\textwidth}{!}{%
\begin{tabular}{llcccccc}
\toprule
Configuration & Steps & p50 ms & p90 ms & mean ms & peak mem & vs 7B vLLM & vs 7B HF (merged) \\
\midrule
Real steps, fp32 (precision of all accuracy tables) & shared 100 & \textbf{74.3} & 127.8 & 86.9 & 3.45\,GB & 2.8$\times$ & 6.8$\times$ \\
Real steps, bf16 autocast & shared 100 & \textbf{51.5} & 79.7 & 56.2 & 2.48\,GB & 4.0$\times$ & 9.8$\times$ \\
Real steps, fp32 & all 200 & 74.8 & 134.0 & 87.5 & 3.45\,GB & -- & -- \\
Real steps, bf16 autocast & all 200 & 48.0 & 79.0 & 53.4 & 2.48\,GB & -- & -- \\
\midrule
Synthetic fixed $N$, fp32, $N{=}5$ / 23 / 57 / 100 & -- & 28.6 / 53.9 / 95.8 / 151.5 & -- & -- & 1.9--3.3\,GB & -- & -- \\
Synthetic fixed $N$, bf16, $N{=}5$ / 10 / 23 / 57 / 100 & -- & 19.4 / 22.2 / 28.3 / 43.6 / 63.8 & -- & -- & 1.8--2.4\,GB & -- & -- \\
\bottomrule
\end{tabular}}
\caption{\textbf{Real-step serving path of the deployed \lwm{} pipeline and the synthetic micro-benchmark} (one NVIDIA H20, batch 1). Real-step rows: the deployed pipeline, timed end to end on real \achigh{} IDD test steps; ``shared 100'' are the steps on which the 7B agent was also timed (identical screenshots, candidates and history), a subset of the ``all 200'' steps timed for \lwm{}. Ratios use the shared 100 steps only, against the same-protocol 7B SFT agent (merged LoRA) timed end to end: 206.7\,ms with vLLM ($\ge$16\,GB peak memory) and 506.7\,ms with HF transformers (15.9\,GB). Synthetic rows: random image tensors and candidate vectors at a fixed number $N$ of scored actions, 100-iteration median, a scaling illustration only. Measurement boundaries: see the common protocol below.}
\label{tab:realstep}
\end{table*}

\begin{figure*}[t]
  \centering
  \includegraphics[width=\textwidth]{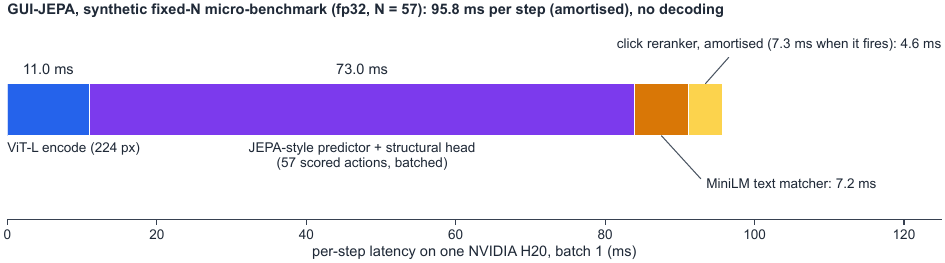}
\caption{\textbf{Latency decomposition in the synthetic fixed-$N$ micro-benchmark} (one NVIDIA H20, batch 1, fp32 eager, $N{=}57$ scored actions, random image tensors and candidate vectors, 100-iteration median). The click-reranker segment is amortised over the action mix; a click step takes 98.4\,ms. The benchmark excludes JPEG decoding, CPU ROI pooling and the span ranker and is a scaling illustration only; real-step medians are in \cref{tab:realstep}.}
\label{fig:latency}
\end{figure*}

\paragraph{Common protocol.}
All latencies are measured on one NVIDIA H20 at batch 1 with a single request and no concurrent load on the GPU, with CUDA synchronisation before and after each timed step, and reported as medians. \lwm{} timing includes JPEG decoding and CPU candidate and ROI preparation, and 7B timing includes image preprocessing and tokenisation; screenshot capture, accessibility-tree retrieval and parsing, and action execution on the device are excluded for every system. Every latency and speed ratio in the main text comes from the 100 real IDD steps on which both systems were timed (\cref{tab:latency,tab:realstep}). The 200-step medians include steps the 7B agent never ran and are given for reference only.

\paragraph{7B agent on real steps.}
The Qwen2.5-VL-7B agent is timed on the first 110 \achigh{} IDD test steps in test-set order (10 warm-up, 100 measured) with the prompt of its accuracy evaluation: the screenshot at a 377-token image budget, the goal, the ground-truth history and the enumerated candidate list in the Qwen chat template, 931 input tokens on average. It decodes greedily until the end-of-sequence token and emits 11.75 new tokens on average, its most favourable configuration. End to end, including CPU prompt preparation (chat template, image processor, tokenisation), HF transformers 4.57 in bf16 with flash-attention~2 needs 506.7\,ms per step for the merged LoRA-SFT model (generation alone 499.1\,ms; prefill 189\,ms, decode 28.3\,ms per token), 504.6\,ms for the zero-shot base model (p90 580.6; prefill 187\,ms, 27.5\,ms per token; 15.9\,GB peak) and 753.6\,ms for the unmerged LoRA model used in the accuracy evaluation (p90 889.3; prefill 229\,ms, 47.8\,ms per token; 16.2\,GB). Merging is the deployment configuration, so our ratios use the merged model. With vLLM 0.11.2 \citep{DBLP:conf/sosp/KwonLZ0ZY0ZS23} in bf16 the merged model needs 206.7\,ms end to end (generation alone 206.5\,ms; prefill 96\,ms, 10.2\,ms per token) with prefix caching off, and 211.9\,ms (p90 250) with prefix caching on, where the 15\,ms prefill and 18.5\,ms per token reflect the cached system prefix.

\paragraph{\lwm{} real-step serving path.}
We time the final pipeline end to end on the first 210 \achigh{} IDD test steps in test-set order (10 warm-up, 200 measured, 46 episodes), so the 7B agent's 100 measured steps are a subset of ours. Each step executes the full pipeline, with the span ranker only on steps predicted as \texttt{input\_text} (10\% of steps) and the click reranker only on steps predicted as click (63.5\%). The 200 steps have 24.4 a11y candidates on average (median 18, p90 49) and 33.0 scored actions (median 28.5, p90 58). The served policy is the evaluated policy: its base-policy action matches the offline evaluation, which uses precomputed encoder features, on 198 of the 200 steps (99.0\%), and on all 200 when both receive the same preprocessed frames. On the 200 steps the bf16 and fp32 pipelines make identical predictions, including the typed text. On all IDD steps, the base policy (base stage of \cref{tab:waterfall}) changes by $+0.18$ Low / $-0.07$ High in bf16 relative to fp32, with 99.8\% step-level agreement on type, candidate and direction. The typed string agrees on 151 of the 322 High steps where both predict \texttt{input\_text}; that ranker is rarely right in High, and High \texttt{input\_text} accuracy moves from 15.6 to 14.2 (4 of 282 steps).

\paragraph{Where the \lwm{} step time goes.}
Over the 200 steps, the fp32 p50 decomposes into 8.3\,ms JPEG decoding and resizing (9.6\% of the mean), 11.4\,ms ViT-L (13.0\%), 1.2\,ms candidate construction, 1.2\,ms ROI encoding, 39.7\,ms predictor and structural head (51.4\%), 8.8\,ms matcher (12.7\%), 8.1\,ms click reranker when it fires (6.8\%) and 21.5\,ms span ranker when it fires (3.3\%); in bf16 the p50 components are 7.0\,ms decoding, 7.7\,ms ViT-L, 11.8\,ms predictor and head and 10.4\,ms matcher. By a11y candidate count the fp32 p50 is 58.9\,ms below 15 candidates, 74.5 at 15--29, 125.1 at 30--59 and 168.7 at 60 or more. The CLIP goal embedding runs once per episode and is not included: its median is 5.6\,ms (mean 10.8\,ms, raised by one 245\,ms first call), about 1.3\,ms per step amortised at the median and 2.4\,ms at the mean. In \aclow{} the step instruction changes at every step, so the CLIP call recurs per step; all real-step timings are \achigh{}.

\begin{figure*}[t]
  \centering
  \includegraphics[width=\textwidth]{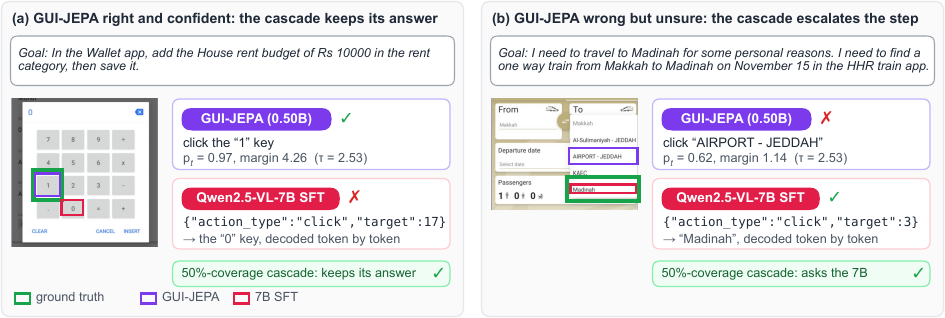}
\caption{\textbf{Complementary errors behind the cascade} on two real \achigh{} IDD steps (\lwm{} with both rankers and the same-protocol 7B SFT agent, given the same candidates). (a)~\lwm{} is right with a margin above the 50\%-coverage threshold $\tau$ (the median margin), so the cascade keeps its answer, while the 7B agent presses the wrong key. (b)~\lwm{} picks the wrong destination with a margin below $\tau$, so the cascade escalates the step to the 7B agent, which is right. Boxes: ground truth (green), \lwm{} (purple), 7B (rose).}
\label{fig:qual}
\end{figure*}

\begin{figure}[t]
  \centering
  \includegraphics[width=\columnwidth]{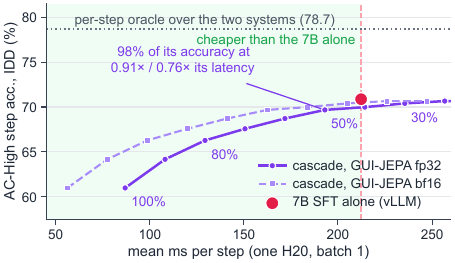}
  \caption{\textbf{\lwm{} as the cheap first stage of a cascade} (\achigh{}, IDD). Steps with a large top-two margin are answered by \lwm{}, the rest by the 7B SFT agent; labels: share of steps kept by \lwm{}. Latency: the mean \lwm{} step plus, on escalated steps, the mean vLLM 7B step (same 100 real steps).}
  \label{fig:routing}
\end{figure}

\paragraph{\lwm{} synthetic fixed-$N$ micro-benchmark.}
The synthetic rows of \cref{tab:realstep} and the lines of \cref{fig:latency_sweep} time the final pipeline in eager PyTorch on a random 224-px image tensor with random action and ROI vectors and dummy candidate texts and bounds, 100-iteration median. $N$ counts the scored actions in the predictor batch, that is, a11y candidates plus global actions. $N{=}23$ and 57 equal the mean and p95 a11y candidate counts of IDD, but a real step scores more actions than it has a11y candidates (33.0 against 24.4 on average on the timed steps), so the synthetic $N{=}23$ point understates a typical step and the real-step rows are the operating point. The encoder cost is fixed, and the predictor batch grows by about 1.2\,ms per scored action in fp32 and 0.37\,ms in bf16; \texttt{torch.compile} on top of bf16 lowers the bf16 times by a further 0.7--6.6\%. The benchmark's inputs are not real screenshots, so it only shows the linear scaling in $N$.

\paragraph{Literature latencies.}
8B-class GUI agents average 2.6--10.6\,s per step in a live-device benchmark (\citealp[Tab.~2]{sui2026androiddailyverifiablebenchmarkmobile}; means over valid sessions, BF16 on H20 GPUs with the GPU count and serving engine not stated, each model under its official inference protocol where one is released). V-Droid reports 0.7\,s per decision, covering all candidates of a step (0.44\,s batched prefill verification plus 0.30\,s action completion), and 4.3\,s per step end to end, 3.03\,s of it in a GPT-4 working-memory update (verifier on two RTX 4090 GPUs; \citealp{dai2026v}). Neither shares our measurement boundary or statistic (means over live-device sessions with model-specific prompts and outputs, against our median over 100 real test steps).

\section{Confidence routing to the 7B agent}
\label{app:routing}

\lwm{} and the same-protocol 7B agent err on different steps (\cref{fig:qual}): on IDD \achigh{} at least one of them is right on 78.9\% of steps on average over the three \lwm{} seeds (78.7--79.2), against 70.87 for the 7B agent and 60.12 for \lwm{}. We therefore test \lwm{} as the cheap first stage of a cascade: a step is answered by \lwm{} when the margin between its top two final scores is at least $\tau$ and escalated to the 7B SFT agent otherwise. The margin predicts whether \lwm{} is correct with AUROC 0.74--0.77, and on its most confident half of the steps \lwm{} alone is 77.6--79.2\% accurate. Setting $\tau$ at coverage quantiles (the share of steps kept by \lwm{}) traces the curve of \cref{fig:routing} (main-text model). At 50\% coverage the cascade reaches 69.48 High on average over seeds, 98\% of the 7B agent's 70.87, at 0.91$\times$ its mean vLLM latency with fp32 \lwm{} and 0.76$\times$ with bf16 (with HF serving, $86.9 + 0.5\times515.7 \approx 345$\,ms against 515.7\,ms, 0.67$\times$). Because $\tau$ is set on coverage quantiles of the same IDD steps, we report the cascade as a deployment mode. For the main-text model the margin AUROC is 0.74, at least one system is right on 78.7\% of steps, and the 50\%-coverage cascade reaches 69.67 (98\% of 70.87) at the same expected latency, with \lwm{} 78.1\% accurate on the steps it keeps.

\end{document}